\documentclass[twoside]{article}

\usepackage[accepted]{aistats2026}
\usepackage[round]{natbib}
\usepackage{comment}
\usepackage{graphicx} 
\usepackage{subcaption} 
\usepackage{researchpack}
\usepackage{algorithm} 
\usepackage{algpseudocode} 
\usepackage{bbm}
\usepackage{hyperref}
\usepackage[capitalize,nameinlink]{cleveref} 
\usepackage{float}
\usepackage{algorithm}
\usepackage{algpseudocode}
\usepackage[table]{xcolor}
\usepackage{natbib}
\usepackage{multicol}
\usepackage{multirow}
\usepackage{tikz}
\tikzset{
  node/.style={draw, rounded corners=3pt, minimum height=1cm, minimum width=1.8cm, align=center},
  arrow/.style={-{Latex[length=3mm]}, thick},
}
\usepackage{nicefrac}

\newcommand{\gdt}[1]{\textcolor{violet}{}}
\title{Multiclass Local Calibration with the Jensen-Shannon Distance}

\newcommand{\LCN}{\textsc{LN}}

\newcommand{\TS}{\textsc{TS}}
\newcommand{\DC}{\textsc{DC}}
\newcommand{\IR}{\textsc{IR}}
\newcommand{\PS}{\textsc{PS}}
\newcommand{\KC}{\textsc{KC}}
\newcommand{\KO}{\textsc{KO}}

\newcommand{\NC}{\textsc{NC}}

\newcommand{\cifarTen}{\texttt{Cifar10}}

\begin{document}
\twocolumn[

\aistatstitle{Multiclass Local Calibration with the Jensen-Shannon Distance}
\runningauthor{Barbera, Perini, De Toni, Passerini and Pugnana}
\aistatsauthor{
Cesare Barbera$^{\diamond,\ast,\dagger}$ \And
Lorenzo Perini$^\star$ \And
Giovanni De Toni$^\ddagger$ \AND
Andrea Passerini$^\ast$ \And
Andrea Pugnana$^{\diamond,\ast}$
}

\aistatsaddress{$^\ast$University of Trento \And $^\dagger$University of Pisa \And  $^\star$Meta \And $^\ddagger$Fondazione Bruno Kessler } ]

\begin{abstract}
Developing trustworthy Machine Learning (ML) models requires their predicted probabilities to be well-calibrated, meaning they should reflect true-class frequencies. 
Among calibration notions in multiclass classification, strong calibration is the most stringent, as it requires all predicted probabilities to be simultaneously calibrated across all classes. However, existing approaches to multiclass calibration lack a notion of distance among inputs, which makes them vulnerable to proximity bias: predictions in sparse regions of the feature space are systematically miscalibrated. 
In this work, we address this main shortcoming by introducing a local perspective on multiclass calibration. First, we formally define multiclass \textit{local calibration} and establish its relationship with \textit{strong calibration}. Second, we theoretically analyze the pitfalls of existing evaluation metrics when applied to multiclass local calibration. Third, we propose a practical method to enhance \textit{local calibration} in Neural Networks, which enforces alignment between predicted probabilities and local estimates of class frequencies using the Jensen-Shannon distance.~Finally, we empirically validate our approach against existing multiclass calibration techniques.

\end{abstract}

\section{INTRODUCTION}
\begingroup
\renewcommand\thefootnote{$\diamond$}
\footnotetext[0]{Corresponding authors. \texttt{\{name\}.\{surname\}@unitn.it}.}
\addtocounter{footnote}{0}
\endgroup
In many high-stakes applications, Machine Learning (ML) models are expected not only to be accurate but also well-calibrated \citep{DBLP:journals/corr/abs-2308-01222, DBLP:journals/cmpb/SambyalNKB23,DBLP:conf/bigdataconf/CianciGGKMPRR23,DBLP:conf/aaai/RuggieriP25} - i.e., their predicted probabilities should reflect the true empirical frequencies of the corresponding classes.
For instance, in the healthcare context, \citet{van2019calibration} compares two cardiovascular risk prediction models over two million patients from the UK. They show that the better-calibrated model, despite having a lower AUC, avoided overestimation of risk. 

However, existing works have mostly focused on \textit{confidence calibration} \citep{DBLP:conf/nips/Le-CozHA24, DBLP:conf/uai/LuoBBZWXSESP22, DBLP:conf/nips/XiongDKW0XH23,DBLP:journals/ml/FilhoSPSKF23}, which only looks at the top-predicted class to check a model's calibration. 
While in binary classification tasks, a well-calibrated model on the top-predicted class ensures good calibration, this is not true for multiclass classification tasks.

\begin{figure*}[t]
    \centering
        \begin{subfigure}{0.35\linewidth}
        \centering
        \includegraphics[width=\linewidth]{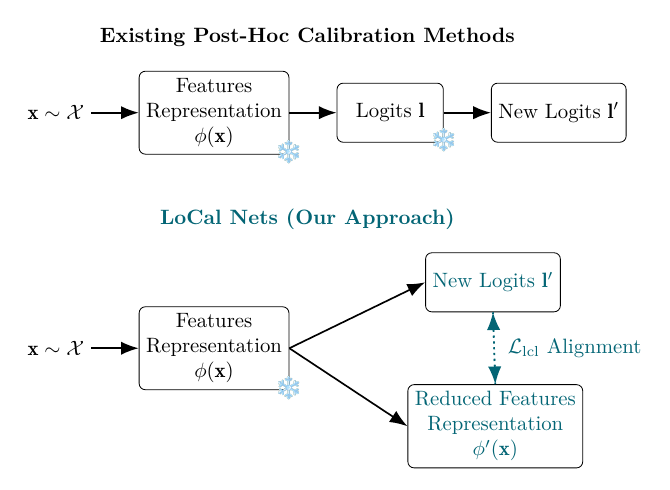}
        \caption{\LCN{} flow}
        \label{fig:arch}
    \end{subfigure}\hfill
    \begin{subfigure}{.63\linewidth}
                \centering
        \includegraphics[width=\linewidth]{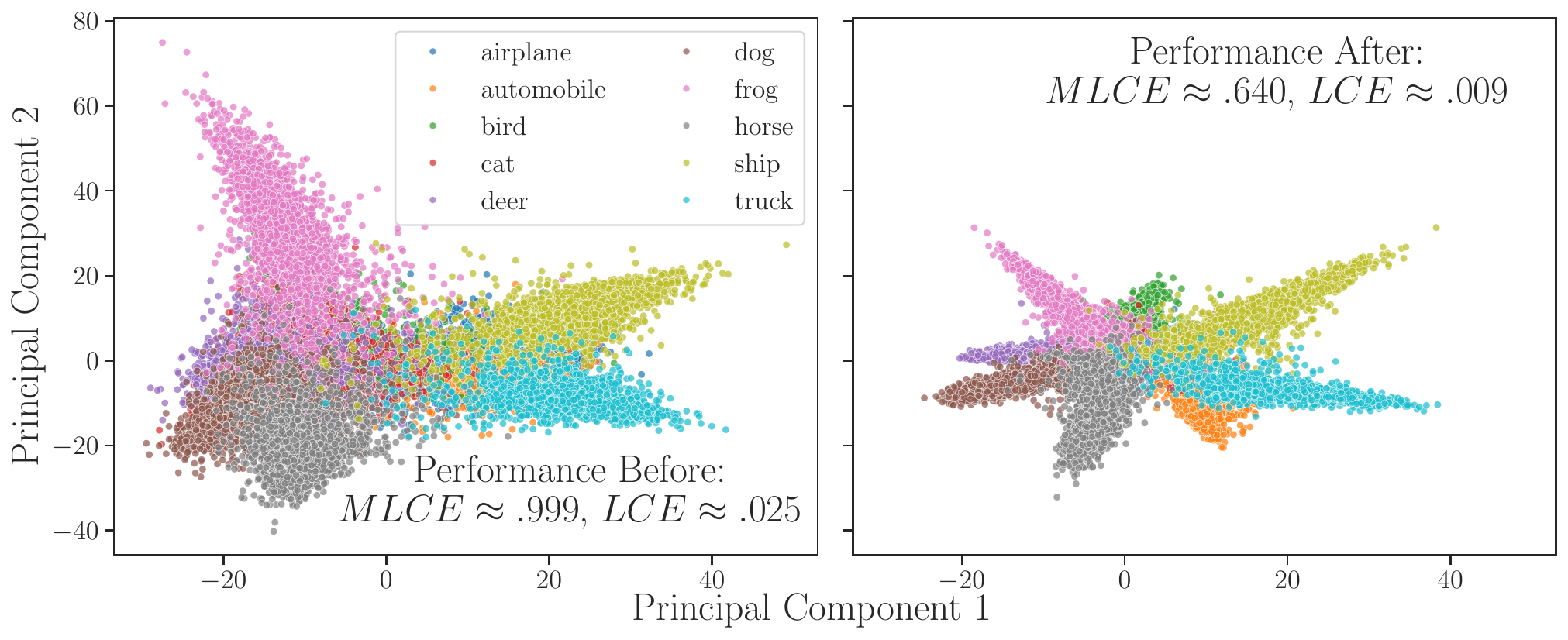}
        \caption{\LCN{} effect on \cifarTen{} using a \texttt{resnet-50}}
        \label{fig:effect}
    \end{subfigure}

    \caption{
        \textbf{Our LoCal Nets (LN) provide local calibration through feature reshaping}
        (\ref{fig:arch}) Unlike post-hoc calibrators that rescale frozen 
        ( \hspace{-.35em}\raisebox{-0.1\baselineskip}{%
        \includegraphics[scale=.2]{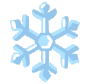}%
        }\hspace{-.35em} )
        logits, LNs jointly (i) learn reduced feature representations $\phi'(\mbx)$ and (ii) output new calibrated logits, aligning predictions with local class frequencies via Jensen–Shannon distance. (\ref{fig:effect}) On \cifarTen{} with \texttt{resnet-50}, LNs yield tighter, better-separated class clusters and improved calibration ($\approx64\%$ reduction in MLCE, $~36\%$ reduction in LCE).
    }
    \label{fig:arch_and_2d}
\end{figure*}

\textit{Example.} 
Consider an ML model trained to predict the stage of cancer in a cell. Suppose the model is calibrated only with respect to its most confident prediction, i.e., if it assigns a probability of $p\%$ to early-stage cancer, then about $p\%$ of the instances receiving such a score are indeed early-stage cancer. Although this might appear sufficient for decision-making~\citep{DBLP:conf/icml/GuoPSW17}, inaccurate probability estimates for less frequent classes can be critical: failing to approximate the likelihood of rare transitional states---such as cells halfway between benign and malignant---might hide patterns about tumor progression.

Hence, stronger notions for multiclass calibration have been proposed. One example is \textit{strong calibration}~\citep{DBLP:conf/aistats/VaicenaviciusWA19}, which requires alignment across the full probability vector. 
However, existing multiclass calibration notions do not consider distances among instances, making them prone to \textit{proximity bias}~\citep{DBLP:conf/nips/XiongDKW0XH23}, whereby predictions for instances in sparsely populated regions of the decision space are more likely to be poorly calibrated. 
This reveals a structural limitation of existing multiclass calibration notions and motivates the need for definitions that explicitly incorporate geometric structure.

\textbf{Our Contributions.} In this work, we tackle this shortcoming by framing multiclass calibration from a ``local'' perspective.
More precisely, 
$(i)$ in \cref{sec:defining} we introduce the notion of \textit{Multi-Class Local-Calibration}, which leverages true empirical frequencies in the neighborhoods of inputs to assess model reliability across the decision-space, and we connect it to \textit{strong calibration};
$(ii)$ we provide theoretical insights on existing pitfalls of current evaluation metrics when applied to multiclass local calibration (Sections~\ref{sec:evaluating} -~\ref{sec:error_nn});
$(iii)$ in \cref{sec:lcn} we propose a new neural network calibration method called LoCal Nets~(\LCN{}s), which exploits the Jensen-Shannon distance to align predicted probabilities and local estimates of class frequencies, while enforcing denser feature representations (Figure~\ref{fig:arch_and_2d}); 
$(iv)$ we evaluate our approach against existing competitors (\cref{sec:exp}), showing how our method improves over local calibration metrics and keeps competitive performance at the global level.

\section{RELATED WORK}

\textbf{Global Calibration Notions.} Due to the stringent requirements of strong calibration and difficulty in both evaluation and enforcement, the literature has proposed several relaxed alternatives.
Examples of such relaxations are \textit{confidence calibration}~\citep{DBLP:conf/icml/GuoPSW17} and \textit{top-label calibration}~\citep{DBLP:conf/iclr/GuptaR22}. While the former considers only the model’s maximum predicted confidence, the latter refines the definition of confidence calibration by conditioning the empirical frequencies on both the model confidence and the model prediction. 
\textit{Class-wise calibration} \citep{DBLP:conf/nips/KullPKFSF19} evaluates calibration for each class marginally, by comparing predicted probabilities with empirical frequencies for that target alone. 
\textit{Top-$r$ calibration} \citep{DBLP:conf/iclr/GuptaRAMS021} considers if the true class falls within the top-$r$ predicted labels and if the cumulative confidence over the top-$r$ classes aligns with observed frequencies. 
\textit{Decision calibration}~\citep{DBLP:conf/nips/ZhaoKSME21} defines calibration with respect to a decision-making policy, requiring predicted and empirical distributions to match according to a decision maker. 
Since these notions are global, they emphasize predicted probabilities while overlooking factors such as instance density or spatial position, failing to capture poor calibration for underrepresented groups.

\textbf{Post-hoc Calibration Methods.} When a trained model suffers from poorly calibrated outputs, many available model-agnostic post-hoc techniques can be employed to adjust its predicted probabilities.
Popular methods include binning \citep{DBLP:conf/icml/ZadroznyE01}, isotonic regression \citep{DBLP:conf/kdd/ZadroznyE02}, logistic scaling \citep{platt1999probabilistic}, temperature scaling \citep{DBLP:conf/icml/GuoPSW17} and parametric forms of scaling \citep{DBLP:conf/aistats/KullFF17,DBLP:conf/nips/KullPKFSF19,matrixScaling}.
We refer to \citet{DBLP:journals/ml/FilhoSPSKF23} for a complete overview of different calibration techniques.~However, all these methods treat logits as fixed inputs to be rescaled. In contrast, our approach learns new feature representations that reshape the geometry of the representation space itself.~Thus, these new neighborhoods better reflect true class frequencies and improve local calibration.

\textbf{Local Calibration Notions.}
Most methods group instances with similar predicted confidences, but ignore that similar scores can be assigned to points in very different regions of the decision space. 
Hence, researchers focused on approaches that incorporate a notion of distance between inputs.
\citet{DBLP:conf/uai/LuoBBZWXSESP22}, \citet{DBLP:conf/iclr/LinT023}, and \citet{DBLP:journals/tmlr/ValkK23} propose recalibration approaches based on kernel regression.
\citet{DBLP:conf/uai/LuoBBZWXSESP22} focus on confidence calibration, leveraging kernel smoothing to locally adjust predicted confidences. In contrast, \citet{DBLP:conf/iclr/LinT023} and \citet{DBLP:journals/tmlr/ValkK23} extend kernel-based recalibration to the multiclass setting.
A key distinction concerns how feature representations are treated. Both \citet{DBLP:conf/uai/LuoBBZWXSESP22} and \citet{DBLP:journals/tmlr/ValkK23} operate on fixed representations.
In contrast, \citet{DBLP:conf/iclr/LinT023} learn new feature representations tailored to kernel-based estimates.
Our approach is closest in spirit to the latter, yet it differs fundamentally in its objective. Rather than using kernel regression as the final predictor, we use kernel-based local frequency estimates only as a supervision signal to induce locality in the model’s probabilistic output.
Moreover, these approaches do not explicitly formalize or evaluate multiclass calibration as a local property. In contrast, our work introduces a formal notion of multiclass local calibration and directly studies its theoretical and empirical implications.
Also \cite{DBLP:conf/nips/MarxZE23} investigates the utility of kernel regressions for calibration, but their proposal directly applies to a model training process. As such, this approach can suffer from computational and data limitations.
Finally, ~\citet{DBLP:conf/iclr/Perez-LebelMV23} shows that whenever the classifier's decision boundary is complex, it can lead to poor calibration of the scores predicted for less likely instances. They introduce (i) the \textit{cancellation effect}, where miscalibration errors within a confidence group offset one another, and (ii) \textit{proximity bias}, reflecting disparities in calibration quality for instances in sparsely populated regions of the decision space. 
Our approach differs as they focus on (i) data density rather than on the position of points in the decision space; (ii) their proposal only addresses the confidence score of the predicted class. We provide in \cref{sec:loc_and_prox} a detailed analysis of the role of \textit{proximity bias} in our proposal and an illustrative example of the pitfalls of density-based recalibration. 
%


\textbf{Multicalibration.}
Another notion, known as \textit{Multicalibration}, requires the model to be calibrated not just on the entire population, but on virtually any sub-population identifiable by a specific hypothesis class~\citep{hebert2018multicalibration,jung2021moment,baldeschi2025multicalibration,jin2025discretization,perini2025mcgrad}.
Thus, the goal of multicalibration approaches is to ensure that models do \textit{not} make biased predictions on any of these sub-populations~\citep{globus2023multicalibrated,noarov2023statistical}. 

While related, multicalibration differs from our setting in two key aspects. First, it is mostly studied for binary/regression tasks, whereas we address \emph{multiclass} calibration and the complexities of the probability simplex. Second, it identifies semantically meaningful subgroups (\textit{e.g.}, \texttt{User Age > 18}) using expressive features in tabular data. 
In contrast, our approach operates in the high-dimensional representation space of neural networks, where dimensions are abstract and ``auditing'' for subgroups via standard multicalibration techniques is computationally intractable.

\section{BACKGROUND}

In this section, we formally introduce the main definitions and metrics used to evaluate calibration. 

\textbf{Definitions.} Let us consider a multi-class classification setting, where $\mcX\subseteq\mathbb{R}^m$ is the feature space and $\mcY = \{0,\ldots, C-1\}$ is a finite target space with $C$ distinct labels. Let us assume we have access to a given dataset \( D = \{(\mbx_i, y_i)\}_{i=1}^n \) of input-output pairs drawn from an unknown joint distribution \( \mathcal{P} \) over \( \mathcal{X} \times \mathcal{Y} \). Each input \( \mbx_i \in \mcX \) is a feature vector of $m$ dimensions, and each label \( y_i \in \mcY \) has a corresponding one-hot encoded vector $\mathbf{y}_i$ indicating the correct class among the \( C \) possible classes. 
We consider a probabilistic classifier \( f \colon \mathcal{X} \rightarrow \Delta^C \), where \( \Delta^C \) is the \( (C - 1) \)-dimensional probability simplex. In words, a probabilistic classifier maps an input \( \mbx \) to a probability distribution over classes, i.e.,  \( f(\mbx) = \mathbf{\hat{p}} \in \Delta^C \), where each entry \( \mathbf{\hat{p}}_{k} = f_k(\mbx) \) of the predicted probability vector $\mathbf{\hat{p}}$ denotes the predicted probability of class \( k \).

In the multiclass context, the weakest notion of calibration is \textit{confidence calibration}~\citep{DBLP:conf/icml/GuoPSW17}, which requires that only the top-predicted probability $\max_{y\in\mcY}\hat{\mbp}_y$ matches the frequency of correct predictions of the classifier.
A more stringent notion is \textit{strong calibration}~\citep{DBLP:conf/aistats/VaicenaviciusWA19}, which requires the target class conditional distribution on any prediction of the classifier to match that prediction, i.e.,:
\[
\mathbb{P}(\mathbf{y}_k = 1 \mid \mathbf{\hat{p}}) =  \mathbf{\hat{p}}_k \quad \text{for all } k \in \{1, \ldots, C\}.
\]

\textbf{Metrics.} Because of the large number of calibration notions, various metrics have been proposed in the literature to evaluate the calibration of classifiers.

One of the most popular metrics is \textit{Expected Calibration Error ($ECE$)}~\citep{DBLP:conf/aaai/NaeiniCH15}, which measures the calibration of a binary classifier. It bins predicted confidence scores into $B$ intervals and compares the average predicted confidence with the empirical accuracy in each bin. 
A multiclass extension of $ECE$ is
\textit{Class-wise ECE}, which calculates ECE separately for each class $c$ and then averages the results.
\textit{Multidimensional Expected Calibration Error} ($MECE$) extends $ECE$ and its class-wise version to multi-class, by using a multidimensional grid binning.
Unfortunately, the combinatorial nature of such a binning hinders its practical application.
\textit{Expected Cumulative Calibration Error (ECCE)} \citep{DBLP:journals/jmlr/IbarraGTTX22} evaluates the cumulative discrepancy between confidence and accuracy across bins rather than averaging them. 

\textit{Local Calibration Error ($LCE$)} quantifies directly the state of \textit{local calibration} of a probabilistic classifier. We provide here the multiclass extension from the original definition by~\citet{DBLP:conf/uai/LuoBBZWXSESP22}:
    \begin{equation*}
    LCE = \frac1{C}\sum_{b=1}^{m_B} 
    \frac{1}{n}\sum_{i\in B_b}\left\|\frac{\sum_{j\in B_b} \bigl(\hat{\mathbf{p}}_{j} - \mathbf{y}_j \bigr)\, k_\gamma(\mbx_i, \mbx_j)}
     {\sum_{j \in B_b} k_\gamma(\mbx_i, \mbx_j)}\right\|_1,
    \end{equation*}
    where~$\mid\mid\cdot\mid\mid_1$ is an appropriate $\ell^1$ norm, $m_b$ the number of used bins, $B_b$ is the set of instances in the $b$-th bin and $k_\gamma(\mbx_i, \mbx_j)$ is a kernel function that weights the influence of neighbouring points of the anchor $\mbx_i$ to its individual $LCE$ score. Its appearance in the denominator as a normalization term ensures weights sum to 1 for every anchor. In practice, this metric captures the differences in the predicted probabilities and the corresponding ground truths for neighbors of an anchor point $\mbx_i$. The shown formalization assumes balanced classes for simplicity and is presented in the most general form leveraging multidimensional binning. In practice, a trivial class-wise adaptation can be implemented. 
    Finally, the \textit{Maximum Local Calibration Error} is
    \(
    MLCE = \max_{i \in D }\left\|\frac{\sum_{j\in b} \bigl(\hat{\mathbf{p}}_{j} - \mathbf{y}_j \bigr)\, k_\gamma(\mbx_i, \mbx_j)}
     {\sum_{j \in b} k_\gamma(\mbx_i, \mbx_j)}\right\|_1.
    \) 
    Notice that, in its original formulation by~\citet{DBLP:conf/uai/LuoBBZWXSESP22}, $LCE$ is limited to confidence calibration, making it insufficient to assess multiclass \textit{local calibration}.

\section{DEFINING MULTICLASS LOCAL CALIBRATION}
\label{sec:defining}
\textit{Local Calibration} requires introducing the concept of distance between samples to capture how one sample’s probabilistic outputs may influence or relate to another’s. Roughly speaking, our definition is based on the intuition that nearby instances should have more similar label distributions and affect each other’s calibration more strongly.~More precisely, we require the model’s predicted probabilities for each sample to be consistent with the locally averaged estimates of the classes' distribution. The degree of \textit{locality} depends on a kernel function $k$ and its bandwidth parameter $\gamma$, which controls the influence of neighboring points on the estimates for each instance.
\begin{definition}[Multiclass Local Calibration]
\label{def:multiclass_local_calibration}
For each instance \( i \in D \), let \( k_\gamma(\mbx_i, \mbx_j) \), with $j \neq i \in D$ be a kernel function with bandwidth $\gamma$. We consider bandwidth sequences $\{\gamma_n\}$ satisfying standard consistency conditions. Define the associated kernel estimator: 
\[
\hat{\theta}(\mathbf y_i \mid \mbx_i) = \frac{\sum_{j \in D} W_j(\mbx_i)\mathbf y_j}{\sum_{j \in D} W_j(\mbx_i)}, \; \quad \;W_j(\mbx_i) \propto k_{\gamma}(\mbx_i, \mbx_j),
\]
where the weights are normalized to sum to $1$. 
Then, \( f \) is \textbf{locally calibrated} on \( D \) if, for all \( \mbx_i \in D \), the predicted probability vector \( \hat{\mathbf{p}}_i = f(\mbx_i) \) is close to the kernel estimate up to a tolerance \( \varepsilon \geq 0 \), i.e.,
\[
\left\| \mathbf{\hat{p}}_i - \hat{\theta}(\mathbf{y}_i \mid \mbx_i) \right\|_1 \leq \varepsilon, \quad \forall i \in \{1, \dots, n\}
\]
Moreover, when $\varepsilon=0$, we say the classifier is \textbf{perfectly locally calibrated}.
\end{definition}  
\noindent  
Please refer to \cref{appendix:conditioning_ass} for a comprehensive description of the \textit{consistency conditions}.
We stress that \cref{def:multiclass_local_calibration} is inherently finite-sample and bandwidth-dependent as it requires alignment with the statistically optimal local frequency estimate at the chosen resolution $\gamma$, not with the true conditional distribution. The consistency assumptions on the kernel estimator relate this empirical notion to population-level calibration measures. 
Notably, the notion of multiclass local calibration is structurally related to strong calibration. On the one hand, strong calibration enforces the alignment of predicted probabilities with empirical frequencies conditional on the predicted probability vector. On the other hand, local calibration constrains this alignment to be pointwise in the feature space. 
As we show in \cref{thm:MECEisboundedunderLocalCal}, satisfying local calibration is asymptotically sufficient to guarantee strong calibration. More precisely, we establish an upper bound on the continuous Multidimensional Expected Calibration Error (MECE). We focus on this formulation since, unlike its binned counterparts, it integrates over the probability simplex and provides a distribution-level characterization of calibration. 
\begin{theorem}[Continuous MECE under Local Calibration]
Define the continuous Multidimensional Expected Calibration Error (MECE) as:
\[
M(f) = \sum_{y\in \mcY} p(y) \, \mathbb{E}_{\mbx \sim p(\mbx)} \left[ \left| \mathbb{E}[\mathbf{y} \mid \mbx]_y - f_y(\mbx) \right| \right], \notag
\]

If a model \( f \) satisfies local calibration, under standard regularity conditions there exists $k \in (0,1]$ such that  continuous $MECE$ is asymptotically upper bounded:
\[
M(f) \leq \varepsilon \cdot k
\]
\label{thm:MECEisboundedunderLocalCal}
\end{theorem} 
%
Thus, in the limit, local calibration can be interpreted as a geometric refinement of strong calibration, ensuring that global reliability emerges from consistent local behavior. A detailed proof is in \cref{sec:proof_thm1}.

\section{EVALUATING MULTICLASS CALIBRATION}
\label{sec:evaluating}
Although the continuous MECE allows for the derivation of theoretical results, such a metric is not computationally feasible in practice. We address this limitation by theoretically analyzing practical methods to evaluate the \textit{local calibration} of probabilistic classifiers.

\begin{definition}[General binning calibration metric]
Let 
\(
\beta:\Delta^C \rightarrow \{1,\ldots,m_B\}
\) be a deterministic binning function that
partitions the probability simplex \(\Delta^C\) into 
\( m_B \) disjoint bins \( \{B_b\}_{b=1}^{m_B} \) and let $\varphi: [0,1]\times[0,1] \rightarrow \mathbb{R}_{\geq0}$ be a scalar comparator that measures discrepancy between an empirical frequency $\text{freq}_{b,c}$ and a predicted confidence $\text{conf}_{b,c}$ that is Lipschitz in both arguments with constant $L_\varphi$.
The general multiclass bin-based calibration error metric is obtained as:
\[
\mathcal{E}(D;\varphi;\beta) = \sum_{b=1}^{m_B} \sum_{c=1}^C w_{b,c}\cdot\pi_c\cdot\varphi(\text{freq}_{b,c},\text{conf}_{b,c})
\]
where $w_{b,c}$ are  possibly class dependent deterministic bin weights and $\pi_c$ are deterministic class weights. 
\label{def:metric_def}
\end{definition}

This general definition of calibration metrics provides a flexible framework that subsumes all calibration measures that leverage binning, including $ECE$ and $MECE$. 
Interestingly, we can define a probabilistic bound for the value of any calibration metric satisfying \cref{def:metric_def} under \textit{local calibration}. See \cref{appendix:general_assumptions} for a detailed description of the assumptions.
\begin{theorem}[Error decomposition of calibration metrics under Local Calibration]
Let $\mathcal{E}(D;\varphi;\beta)$ be a calibration metric satisfying \cref{def:metric_def}, evaluated on a sample $\{(\mbx_i,\mathbf y_i)\}_{i=1}^n$. 
Let $f$ be locally calibrated with error $\varepsilon$ relative to kernel estimates $\hat \theta(\mathbf y_i \mid \mbx_i)$ computed on an independent arbitrarily large auxiliary dataset. 
Conditioning on the evaluation inputs $\{\mbx_i\}_{i=1}^n$ and the corresponding  consistent kernel estimates, for any $\delta \in (0,1)$, with probability at least $1-\delta$:
\[
\mathcal{E}(D;\varphi;\beta)
\le L_\varphi \sum_{b=1}^{m_B} \sum_{c=1}^C w_{b,c} \pi_c
\Bigg( \sqrt{\frac{\log(\frac{2Cm_B}{\delta})}{2 |\Psi(b,c)|}} \;+\; \varepsilon \Bigg) 
\]
where $\Psi(\cdot, \cdot)$ selects a bin based on the index $b$ and possibly the label $c$.
\label{thm:generic_bound}
\end{theorem}

\noindent  We provide the proof in the \cref{proof:thm2_proof} together with a discussion of the result under limited cardinality of the auxiliary dataset leveraged for kernel estimates. We also show the result extends to cumulative metrics like $ECCE$ in Appendix~\ref{sec:extension_ecce}. 
The bound reveals a natural decomposition of calibration error into two components: a deterministic term proportional to the local calibration error $\varepsilon$, and a stochastic term that arises inevitably from finite-sample estimation of empirical frequencies within bins. Even when a model is perfectly locally calibrated ($\varepsilon=0$), the calibration metric remains subject to sampling fluctuations that vanish only as bin cardinalities grow.
Notice that Theorem \ref{thm:generic_bound} establishes an upper bound on $\mathcal{E}(D;\varphi;\beta)$ in terms of $\varepsilon$, implying that a high value of the metric necessarily indicates poor \textit{local calibration}. However, the converse does not hold in general, as binning-based metrics are known to suffer from \textit{cancellation effects}, whereby opposing errors may offset each other and obscure miscalibration. 
%

\section{EVALUATING CALIBRATION IN NEURAL NETWORKS}
\label{sec:error_nn}
In the following section, we focus on Neural Network classifiers.~Exploiting their structure and the \(L\)-Lipschitz continuity property with respect to the norm \( \|\cdot\|_1 \)  \citep{DBLP:journals/corr/abs-1704-00805}, we refine the general results presented in the previous sections and investigate further the properties of \textit{Local Calibration Error}. Practically speaking, achieving \textit{perfect local calibration} is unfeasible. In the ideal case where \( \varepsilon = 0 \), the distribution of points would be both perfectly centered around \( \mbx_i \) and fully representative of the local data distribution. 
However, such conditions are never met in finite samples. 
Additionally, \cref{def:multiclass_local_calibration} does not take into account potential disparities in the miscalibration error of classes but only bounds the total sum of errors.
For these two reasons, we relax \cref{def:multiclass_local_calibration} by bounding the admissible error in the predicted probability for class $c$ by the maximum change in the output probabilities that can occur when moving within a ball of radius $\rho$ in the feature space.
Thus, decreasing \(\rho\) bounds the class-wise deviation a model can tolerate while approaching \textit{perfect local calibration}. Note that our definition operates under the assumption that the kernel estimator refers to the Neural Network learned feature-representations as in \cite{DBLP:conf/uai/LuoBBZWXSESP22}.
\begin{definition}[$\rho$-Perfect Local Calibration]
\label{def:perfect_local_calibration_features}
Let \(f\) be composed of a feature extractor \( \phi: \mathcal{X} \to \mathcal{F} \) and a classification layer \( g: \mathcal{F} \to \Delta^C \). In addition, \(\phi(\mbx_i) \) is assumed Lipschitz-continuous with respect to the softmax with constant \( L > 0 \). 
We say that \( f \) is $\rho$-perfectly locally calibrated if for every instance \( \mbx_i \in D \) and for every class \( c \in \{1, \ldots, C\} \), the absolute calibration error is bounded as follows:
\[
\left| \mathbf{\hat{p}}_{i,c} - \hat{\theta}_c(\mathbf y \mid \phi(\mbx_i)) \right| \leq L \cdot\rho. 
\]
\end{definition}
%
%
%
We exploit the $\rho$-\textit{perfect local calibration} definition to further improve the \cref{thm:generic_bound} bound. We formalize it in the following corollary:

\begin{corollary}[Calibration measure under $\rho$-Perfect Local Calibration]

If a classifier $f$ satisfies $\rho$-\textit{perfect local calibration}, the error \( \mathcal{E}(D;\varphi;\beta) \) becomes purely stochastic fluctuation. For any \(\delta \in [0,1]\), with probability at least $1-\delta$ it holds that: 
\[
\limsup_{\rho \to 0} \mathcal{E}(D;\varphi;\beta) \leq L_\varphi\sum_{b=1}^{m_B} \sum_{c=1}^C w_{b,c}\pi_c\sqrt{ \frac{ \log(\frac{2Cm_B}{\delta}) }{ 2 |\Psi(b,c)| } }
\]
\end{corollary}
\noindent Proof is in \cref{subsec:proof_cor1}. This result underscores that the smaller the value of $\rho$ for which a model exhibits perfect \textit{local calibration}, the more the overall error is dominated by the stochastic component. 


Our theoretical analysis on the role of \textit{local calibration} in shaping the behavior of calibration metrics will next address the multiclass version of
$LCE$. Its inclusion is essential as it is the only metric that directly quantifies the degree of \textit{local calibration} of a probabilistic classifier.
More precisely, we show that, unlike global calibration measures, LCE relies on kernel regression estimates, which introduces a bias–variance trade-off on top of the inherent calibration error. Roughly speaking, the bound we derive decomposes into three interpretable components:
(i)~a calibration term $\varepsilon$ controlled by the \textit{local calibration} property;
(ii)~a variance term growing as the kernel weights become concentrated on fewer neighbors; and
(iii)~a bias term which penalizes assigning large weights to distant samples. We discuss in detail assumptions in \cref{appendix:general_assumptions}.
 
\begin{theorem}[Probabilistic bound for multiclass LCE under Local Calibration]
\label{thm:lc_bound}
Let \( f\) be composed of a feature extractor \( \phi\) and classification layer \( g\).
Let $f$ be locally calibrated with error $\varepsilon$ relative to kernel estimates $\hat \theta(\mathbf y_i \mid \mbx_i)$ computed on an independent arbitrarily large auxiliary dataset. 
Let the LCE be computed on $\phi(\mbx_i)$. 
Then, conditioning on the evaluation inputs $\{\mbx_i\}_{i=1}^n$ and on the corresponding consistent kernel estimates, for any \(\delta \in [0,1]\), with at least probability  \(1-\delta\): 

\allowdisplaybreaks
\begin{align*}
&\mathrm{LCE} \le \; \varepsilon \;+\; \overbrace{\frac{2}{n}\sum_{b=1}^{m_B}\sum_{i \in I_b}\sqrt{\frac{\log\!\bigl(\frac{2nC}{\delta}\bigr)}{2n_i^{\mathrm{eff}}}}}^{\text{variance term}} \\ \: + \:
& \underbrace{\frac{L}{n}\sum_{b=1}^{m_B}
\mathbb{E}\!\Big[\sum_{i\in I_b}\sum_{j\in I_b} w_{i,j}\|\phi(\mbx_j)-\phi(\mbx_i)\|_1\Big]}_{\text{bias term}}
\end{align*}
where \(
n_i^{\mathrm{eff}}= \frac{1}{\sum_{j \in I_b} w_{i,j}^2}
\), with $w_{i,j}=\frac{k_{\gamma}(\phi(\mbx_i), \phi(\mbx_j))}{\sum_j k_{\gamma}(\phi(\mbx_i),\phi(\mbx_j))}$.

\end{theorem}
We provide the proof in \cref{sec:proof_thm3}. 
This decomposition highlights the fundamental bias–variance trade-off induced by kernel smoothing:~tighter kernels reduce bias at the expense of increased variance, while broader kernels reduce variance but incur larger bias. Therefore, estimating the order of magnitude of these terms clarifies the state of \textit{local calibration} for a model.

In conclusion, the results presented demonstrate that both binning-based metrics and kernel-based metrics like LCE admit a probabilistic decomposition under the assumption of \textit{local calibration}. 
This theoretical perspective clarifies the role of \textit{local calibration} in shaping the behavior of multiclass calibration metrics and highlights practical limitations of the metrics themselves in capturing the phenomenon. 

\section{IMPROVING LOCAL CALIBRATION IN PRACTICE}
\label{sec:lcn}

We now focus on improving the \textit{local calibration} of Neural Networks in practice. We first introduce \textit{Local Calibration Networks} (\LCN{}), i.e., a two-component neural network architecture, designed to produce representations that exhibit improved \textit{local calibration}. 

Figure~\ref{fig:arch} shows the structure of \LCN{}. While standard post hoc calibrators consider the features representation as fixed, one component of \LCN{}s produces a new reduced-dimensionality feature representation $\phi'(\mbx_i)$ and the other component parametrizes new logits $\mathbf{l}'$. From this perspective \LCN{}s can be viewed as a representation-level post-hoc calibration method.
The two components are trained to minimize the misalignment between the probabilistic outputs and the local estimates of the class distribution computed from \(\phi'(\mbx_i)\). To obtain these estimates, one can resort to kernel-based methods, e.g., Nadaraya-Watson estimators \citep{nadaraya1964estimating, watson1964smooth}, which require specifying a bandwidth hyper-parameter $\gamma$. 
More precisely, \LCN{}s minimize the following loss: 
\begin{align}
\mathcal{L}_{\text{LN}}(\mbx_i,\mathbf{y}_i) = \frac{1}{n} \sum_{i=1}^n \overbrace{\text{d}_{\mathrm{JSD}}\left( \mathbf{\hat{p}}_i , \hat{\theta}(\mathbf{y}_i\mid \phi'(\mbx_i)\right)}^{\text{Alignment term}} + \notag\\\lambda \cdot \underbrace{\mathcal{L}_{\text{ce}}\left( \mathbf{y}_i, \hat{\theta}(\mathbf{y}_i\mid \phi'(\mbx_i )\right) }_{\text{Similarity term}},
\end{align}
where \(\text{d}_{JSD}(P, Q)\) is the Jensen-Shannon distance\footnote{$\text{d}_{JSD}(P, Q) :=\sqrt{\frac1{2}\text{KL}(P\|\frac{P+Q}{2})+\frac1{2}\text{KL}(Q\|\frac{P+Q}{2})}$}\citep{DBLP:conf/isit/FugledeT04, DBLP:journals/tit/EndresS03}  and \(\mathcal{L}_{\text{ce}}\) is the categorical cross entropy. Notice that two distinct terms compose  $\mathcal{L}_{\text{LN}}$, i.e., the \textit{alignment term} and \textit{the similarity term}.

On the one hand, the \textit{alignment term} leverages the Jensen-Shannon distance between the model’s predicted probability vector and the kernel estimates for each instance in the training batch. We prove that, for a consistent estimator, the \textit{alignment term} asymptotically converges to the divergence between the model prediction distribution and the true label distribution. This result relies on \textit{kernel consistency} assumptions described in detail in \cref{appendix:general_assumptions}.
\begin{theorem}[Asymptotic consistency of JSD]
Let $f$ be a probabilistic classifier and let $\hat{\theta}(\cdot \mid \mathbf{x})$ be a kernel estimator of $p(\cdot \mid \mathbf{x})$ that is consistent in mean squared error. Then, under standard regularity conditions:
\begin{align}
    \lim_{n \to \infty} \frac{1}{n} \sum_{i=1}^n \text{d}_{\mathrm{JSD}} \left( \mathbf{\hat{p}}_i, \hat{\theta}(\mathbf y\mid \mbx_i \right) = \notag \\
    \lim_{n \to \infty} \frac{1}{n} \sum_{i=1}^n \text{d}_{\mathrm{JSD}} \left( \mathbf{\hat{p}}_i,\mathbf{p}_i\right).
\end{align}
\label{thm:jsd_cons}
\end{theorem}
We provide proof in the \cref{sec:proof_thm4}. Notably, the \textit{alignment term} allows to learn probabilistic outputs that match the empirical neighbourhood frequencies.

On the other hand, the \textit{similarity term} leverages the categorical cross entropy between the ground truth and the kernel estimates. 
This term serves a similar scope to the one of \cite{DBLP:conf/iclr/LinT023}.
Intuitively, it encourages points with the same label to be attracted to nearby neighbors, but because the kernel is local, distant points of the same class exert little influence on each other. As a result, points that share fine-grained similarities are placed closer together, while more distinct variants remain further apart, but within the same class cluster. We provide a visual intuition of this behavior in Figure~\ref{fig:effect}.
However, the \textit{similarity term} could still collapse class representations if not regularized or if the kernel bandwidth is too wide. Hence, we introduce another regularization hyper-parameter $\lambda$ (to be fine-tuned) to prevent this behaviour.

As a concluding remark, our method leverages kernel estimates during training but does not require them for inference, therefore fully maintaining the efficiency of feed-forward neural networks.

\section{EXPERIMENTAL EVALUATION}
\label{sec:exp}
In this section, we address the following questions:
\begin{itemize}
    \item \textbf{Q1:} Does our proposal outperform baselines on \textit{local calibration} metrics?
    \item \textbf{Q2:} Does our approach match the performance of baselines in global calibration metrics?
    \item \textbf{Q3:} Does our method affect predictive performance?

\end{itemize}
The code\footnote{Code developed by University of Trento authors.} to reproduce our results can be found at
\url{https://github.com/Cesbar99/multiclass-local-calibration}

\subsection{Experimental Settings}

\begin{figure*}[t]
    \centering
    \includegraphics[width=\textwidth]{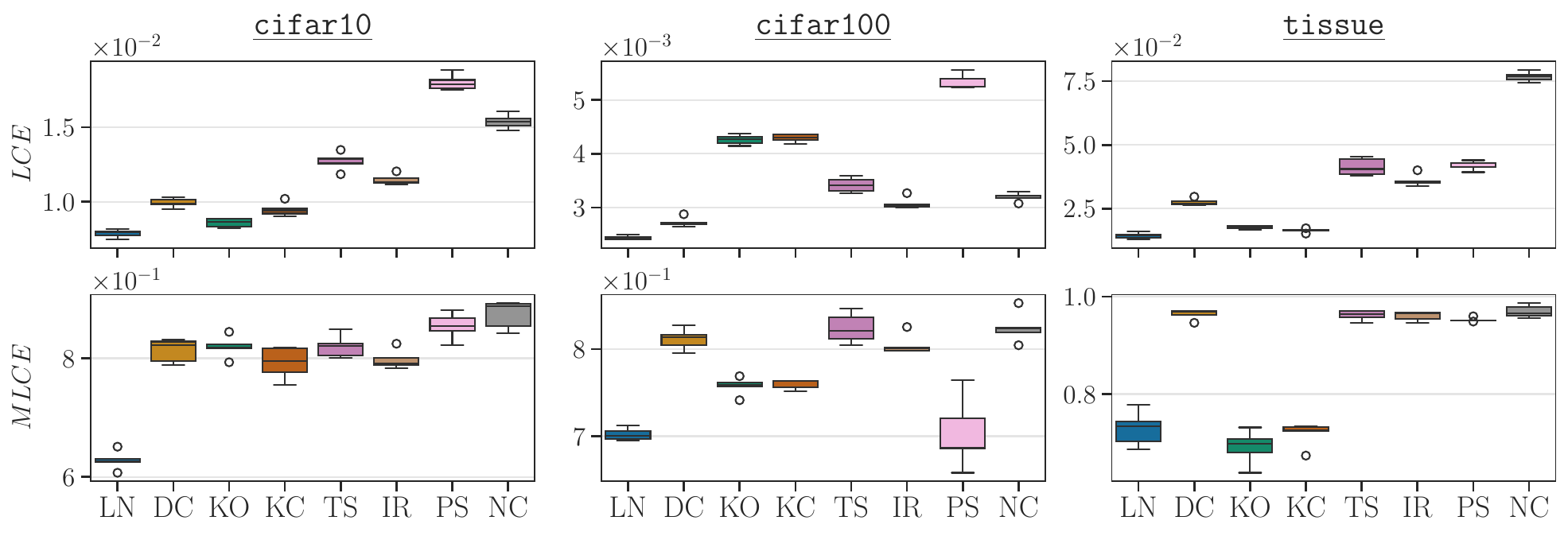}
        \caption{\textbf{Empirical local calibration metrics (\textbf{Q1}) over five runs.} The lower the better.}
    \label{fig:local_metrics}
\end{figure*}

\begin{figure*}[t]
    \centering
    \includegraphics[width=\textwidth]{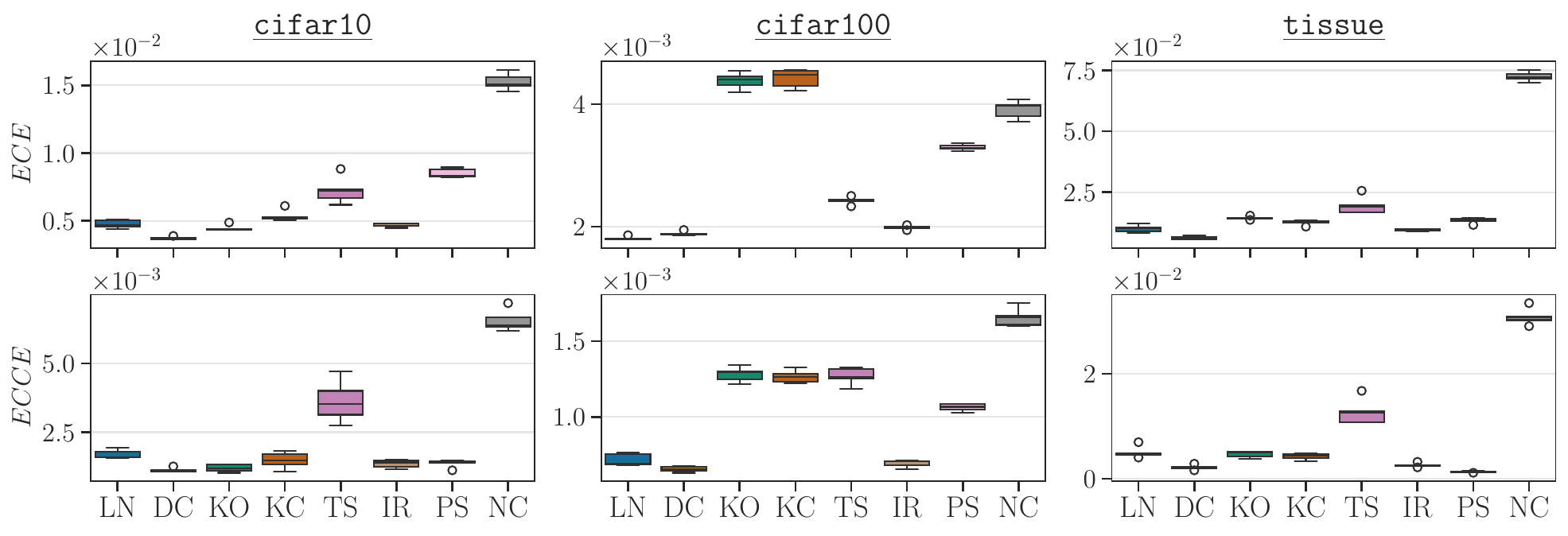}
    \caption{\textbf{Empirical global calibration metrics (\textbf{Q2}) over five runs.} The lower the better.}
    \label{fig:global_metrics}
\end{figure*}

\begin{table*}[t]
\centering
\caption{Results for \textbf{Q3}. We report $Acc$ (the higher the better) and $NLL$ (the lower the better) across datasets. We highlight in bold the best performer and we underscore the second-best method for each dataset and metric.}
\normalsize 
\setlength{\tabcolsep}{6pt} 
\renewcommand{\arraystretch}{1.1} 
\begin{tabular}{c | c c c| c c c}

     \multirow{2}{*}{\textbf{\multirow{2}{*}{\textbf{Method}}}} & \multicolumn{3}{l}{$ACC\,\uparrow$} & \multicolumn{3}{l}{$NLL\,\downarrow$} \\
              \\
              &           \texttt{cifar10} &          \texttt{cifar100} &            \texttt{tissue} &           \texttt{cifar10} &           \texttt{cifar100} &             \texttt{tissue} \\
\midrule
$\textsc{LN}$ & \cellcolor{blue!10}            $.888\pm .002$ & \cellcolor{blue!10}$\underline{.688\pm .001}$ & \cellcolor{blue!10}            $.630\pm .001$ & \cellcolor{blue!10}            $.347\pm .002$ & \cellcolor{blue!10}   $\mathbf{1.125\pm .002}$ & \cellcolor{blue!10}   $\mathbf{1.012\pm .003}$ \\
$\textsc{DC}$ & \cellcolor{gray!10}            $.890\pm .002$ & \cellcolor{gray!10}   $\mathbf{.690\pm .002}$ & \cellcolor{gray!10}            $.617\pm .003$ & \cellcolor{gray!10}   $\mathbf{.332\pm .007}$ & \cellcolor{gray!10}$\underline{1.154\pm .009}$ & \cellcolor{gray!10}            $1.052\pm .009$ \\
$\textsc{KO}$ & $\underline{.893\pm .001}$ &             $.679\pm .003$ & $\underline{.632\pm .001}$ &             $.340\pm .002$ &             $1.352\pm .009$ &             $1.027\pm .002$ \\
$\textsc{KC}$ & \cellcolor{gray!10}   $\mathbf{.893\pm .001}$ & \cellcolor{gray!10}            $.679\pm .004$ & \cellcolor{gray!10}   $\mathbf{.632\pm .001}$ & \cellcolor{gray!10}$\underline{.340\pm .003}$ & \cellcolor{gray!10}            $1.351\pm .007$ & \cellcolor{gray!10}$\underline{1.021\pm .002}$ \\
$\textsc{PS}$ &             $.884\pm .001$ &             $.670\pm .002$ &             $.603\pm .008$ &             $.466\pm .004$ &             $1.618\pm .007$ &             $1.180\pm .008$ \\
$\textsc{IR}$ & \cellcolor{gray!10}            $.884\pm .001$ & \cellcolor{gray!10}            $.670\pm .002$ & \cellcolor{gray!10}            $.603\pm .008$ & \cellcolor{gray!10}            $.364\pm .008$ & \cellcolor{gray!10}            $1.437\pm .029$ & \cellcolor{gray!10}            $1.096\pm .016$ \\
$\textsc{TS}$ &             $.884\pm .001$ &             $.670\pm .002$ &             $.603\pm .008$ &             $.362\pm .008$ &             $1.277\pm .009$ &             $1.112\pm .023$ \\
$\textsc{NC}$ & \cellcolor{gray!10}            $.884\pm .001$ & \cellcolor{gray!10}            $.670\pm .002$ & \cellcolor{gray!10}            $.603\pm .008$ & \cellcolor{gray!10}            $.494\pm .018$ & \cellcolor{gray!10}            $1.502\pm .036$ & \cellcolor{gray!10}            $2.100\pm .101$ \\
\bottomrule
\end{tabular}
\label{tab:Q3res}
\end{table*}
\textbf{Datasets.} We evaluate our research questions over three multiclass datasets, i.e., \texttt{cifar10}, \texttt{cifar100}~\citep{krizhevsky2009learning} and \texttt{tissuemnist} from the \texttt{MedMNIST} collection~\citep{DBLP:conf/isbi/YangSN21}.

\textbf{Methods and Architectures.}
We evaluate our approach, \LCN{}, against different calibration baselines, including Temperature Scaling (\TS{})~\citep{DBLP:conf/icml/GuoPSW17}, Isotonic Regression (\IR{})~\citep{DBLP:conf/kdd/ZadroznyE02}, and Platt Scaling (\PS{})~\citep{platt1999probabilistic}. In addition, we compare with Dirichlet Calibration (\DC{})~\citep{DBLP:conf/nips/KullPKFSF19}, the current state-of-the-art for multiclass calibration. 
Moreover, we consider two K-Cal implementations~\citep{DBLP:conf/iclr/LinT023}: the first one considers the kernel classifier trained using the same loss as \LCN{}~(\KO{}); the second one does not use the extra Jensen-Shannon regularization as in the original version (\KC{}). We also report the performance of the original non-calibrated classifier (\NC{}). 

All methods are based on a \texttt{ResNet-50} backbone for \texttt{CIFAR-10} and \texttt{TissueMNIST}, and a \texttt{ResNet-152} for \texttt{CIFAR-100}. 
We detail hyperparameters in Appendix.

\textbf{Metrics.}
For \textbf{Q1}, we compute two \textit{local calibration} metrics, i.e., class-wise $LCE$ and $MLCE$.
For \textbf{Q2}, we consider two global calibration metrics, i.e., $ECE$ and $ECCE$. 
For \textbf{Q3}, we report two performance measures, i.e., accuracy ($Acc$) and Negative Log-Likelihood ($NLL$). Implementation details are provided in \cref{sec:metrics_impl}.

\textbf{Experimental Setup.}
To evaluate calibration for all methods, we separate the data into three disjoint parts: $(i)$ a training set (further split into training and validation); $(ii)$ a calibration set (with an internal calibration/validation split); and $(iii)$ a held-out test set used exclusively for evaluation. 

For \texttt{CIFAR-10} and \texttt{CIFAR-100}, we use 45\% of the original training data for training, 10\% for validation, and 45\% as test data, with the original test split serving as calibration data. 

For \texttt{TissueMNIST}, we use the pre-computed splits, but since calibration metrics computation requires larger sample sizes to be meaningful, we split the training data in half, yielding balanced training and test splits ($\approx82.5k$ each). The remaining pre-computed test set ($\approx40k$ instances) is used for calibration.

For all the datasets, we split the calibration data into two sets, one to learn the calibration technique ($90\%$ of the calibration data) and one for validation (the remaining $10\%$ of the calibration data).
Then, we $(i)$ train classifiers on the training set; $(ii)$ calibrate using the different methods on the calibration set; $(iii)$ compute the six metrics on the test set. We repeat this procedure using 5 different seeds and average results.
Implementation details are provided in the Appendix.

\subsection{Experimental Results}

\textbf{Q1: Local Nets consistently achieve superior results on local calibration metrics.} Figure~\ref{fig:local_metrics} evaluates the methods in terms of \textit{local calibration}. Across all datasets, \LCN{} consistently emerges as the best-performing approach, providing substantial improvements over competing methods.

On \texttt{cifar10}, \LCN{} achieves the lowest $LCE$ ($.0079 \pm .0003$), followed by \KO{} ($.0086 \pm .0003$), with \KC{} performing at a similar level to \KO{}. In terms of $MLCE$, \LCN{} significantly outperforms all competitors with differences as large as $\approx.16$.

On \texttt{cifar100}, \LCN{} again attains the best $LCE$, 
setting a clear gap with all competitors. Here, \LCN{} achieves $MLCE\approx.7022 \pm .0070$,  with \PS{} nearly matching its result $(MLCE\approx.7031 \pm .0409)$. However, the lower variance of \LCN{} highlights its reliability. The relative competitiveness of \PS{} on this dataset is due to the limited per-class sample size (at most 600 instances), which favors parametric methods such as \PS{}, whereas \LCN{} benefits more from larger sample sizes.

Notably, \LCN{} achieves the best performance in terms of $LCE$ also on \texttt{tissuemnist}, with $LCE\approx.0144 \pm .0012$, outperforming \KC{} ($LCE\approx.0165 \pm .0008$) and all other baselines. In terms of $MLCE$, \KO{} achieves the best performance with $.6913\pm.0349$ while \LCN{} reaches similar results $(MLCE\approx.7293\pm.0359)$. 

Overall, these results demonstrate that \LCN{} provides significant improvements in \textit{local calibration}, particularly in settings with sufficient sample size.

\textbf{Q2: Local Nets achieve competitive global calibration results across datasets.}
Figure~\ref{fig:global_metrics} summarizes the performance of all methods on global calibration metrics. We aim to show that \LCN{} performs at a level comparable to the existing baselines.

For $ECE$, \DC{} is the strongest performer over \texttt{cifar10} with $ECE\approx.0037 \pm .0001$, followed by \KO{} with $ECE\approx.0045 \pm .0002$. Still, \LCN{} reaches comparable performance $(ECE\approx .0048\pm.0003)$. For $ECCE$, \DC{} again leads with $\approx.0011 \pm .0001$, while \LCN{} achieves a level comparable to the other baselines.
On \texttt{cifar100}, \LCN{} achieves the best performance in terms of $ECE$
and is tied with \DC{} and \IR{} on $ECCE$ (Figure~\ref{fig:global_metrics}).
On \texttt{tissuemnist}, \DC{} again performs best with $ECE\approx.0062 \pm .0007$, while \LCN{} ranks third 
(Figure~\ref{fig:global_metrics}). For $ECCE$, \LCN{} performs slightly worse, obtaining $\approx.0050 \pm .0011$.

In summary, \DC{} is the overall best-performing method for global calibration, with its most pronounced advantage on balanced, low-class datasets such as \texttt{cifar10}. Still, our proposed \LCN{} consistently achieves competitive results across datasets.

\textbf{Q3: Beyond calibration, Local Nets enhance predictive performance.} Table~\ref{tab:Q3res} reports results on predictive performance, measured by $NLL$ and $ACC$. Our method achieves the largest reductions in $NLL$ across datasets, with the sole exception of \texttt{cifar10}, where \LCN{}s rank second ($.347 \pm .002$) compared to \DC{} ($.332 \pm .007$). On \texttt{cifar100}, \LCN{} clearly outperforms all competitors with the lowest $NLL$ ($1.125 \pm .002$ vs. $1.265 \pm .016$ for \DC{}). 
A similar trend is observed on \texttt{tissuemnist}, where \LCN{} achieves $1.012 \pm .003$, surpassing \KC{} ($1.021 \pm .002$).

Like all the non-accuracy-preserving methods, we observe gains for \LCN{}: $+0.4\%$ on \texttt{cifar10}, $+1.9\%$ on \texttt{cifar100}, and $+2.7\%$ on \texttt{tissuemnist}. This occurs as other methods are forced to work only on logits, while \LCN{} can learn new feature representations, possibly improving the predictions' quality.

These results highlight that beyond improving calibration, our approach translates into tangible gains in predictive performance, particularly on challenging datasets with larger class cardinality or imbalance.

\section{CONCLUSIONS AND FUTURE WORK}

\textbf{Conclusion.} In this work, we introduced a formal definition of multiclass \textit{local calibration} and theoretically analyzed widely used calibration metrics under this assumption. Building on these insights, we proposed a novel post-hoc calibration method for neural networks that explicitly targets \textit{local calibration} properties. Our empirical evaluation on both benchmarking and real-world datasets demonstrates that the proposed approach yields significant improvements in \textit{local calibration}, while maintaining competitive performance with respect to global calibration metrics. These results highlight the importance of incorporating locality into the design of calibration methods. 

\textbf{Limitations and Future Work.} Although Theorem~\ref{thm:jsd_cons} establishes the consistency of our loss function, this guarantee holds only asymptotically. In finite-sample regimes, kernel estimates may suffer from non-negligible bias, which can limit performance. An important research direction is to explore adaptive kernel choices and scalable training procedures that better manage the bias–variance tradeoff inherent to local estimation. Moreover, the hyperparameter 
$\gamma$ requires careful tuning: small values yield unreliable estimates due to data sparsity, while large values obscure locality. A promising extension is to replace fixed kernels with adaptive or learned similarity functions, potentially improving \textit{local calibration}. Finally, while our method is tailored to neural networks, extending it to enforce \textit{local calibration} across other classes of models remains an important open direction.

\section*{Acknowledgements}

We thank the anonymous reviewers for their feedback.

The research of Cesare Barbera, Giovanni De Toni, Andrea Passerini, and Andrea Pugnana was partially supported by the following projects: Horizon Europe Programme, grants \#10112\-0237-ELIAS and \#101120763-TANGO.
Funded by the European Union. Views and opinions expressed are however those of the author(s) only and do not necessarily reflect those of the European Union or the European Health and Digital Executive Agency (HaDEA). Neither the European Union nor the granting authority can be held responsible for them.
The research of Cesare Barbera, Giovanni De Toni, Andrea Passerini, and Andrea Pugnana was also supported by Ministero delle Imprese e del Made in Italy (IPCEI Cloud DM 27 giugno 2022 – IPCEI-CL-0000007), PNRR-M4C2-Investimento 1.3, Partenariato Esteso PE00000013-“FAIR-Future Artificial Intelligence Research”, funded by the European Commission under the NextGeneration EU programme.

\bibliography{bibliography}
\bibliographystyle{apalike}
\section*{Checklist}



\begin{enumerate}

  \item For all models and algorithms presented, check if you include:
  \begin{enumerate}
    \item A clear description of the mathematical setting, assumptions, algorithm, and/or model. \textbf{[Yes]} We clearly state the settings and assumptions of our method in Section \ref{sec:lcn}.
    \item An analysis of the properties and complexity (time, space, sample size) of any algorithm. \textbf{[Not Applicable]} We do not study these properties in our work.
    \item (Optional) Anonymized source code, with specification of all dependencies, including external libraries.\textbf{ [Yes] } The code to reproduce our results can be found at \url{https://anonymous.4open.science/r/local-calibration-25A3}.
  \end{enumerate}

  \item For any theoretical claim, check if you include:
  \begin{enumerate}
    \item Statements of the full set of assumptions of all theoretical results. \textbf{[Yes]} All our theorems and definitions in Sections \ref{sec:defining}, \ref{sec:evaluating}, \ref{sec:error_nn}, \ref{sec:lcn} clearly state assumptions.
    \item Complete proofs of all theoretical results. \textbf{[Yes]} The proofs can be found in Appendix (Section Proofs)
    \item Clear explanations of any assumptions. [Yes] See Sections \ref{sec:defining}, \ref{sec:evaluating}, \ref{sec:error_nn}, \ref{sec:lcn}.
  \end{enumerate}

  \item For all figures and tables that present empirical results, check if you include:
  \begin{enumerate}
    \item The code, data, and instructions needed to reproduce the main experimental results (either in the supplemental material or as a URL). \textbf{[Yes]} The code to reproduce our results can be found at \url{https://anonymous.4open.science/r/local-calibration-25A3}.
    \item All the training details (e.g., data splits, hyperparameters, how they were chosen). \textbf{[Yes]} See Section \ref{sec:exp}.
    \item A clear definition of the specific measure or statistics and error bars (e.g., with respect to the random seed after running experiments multiple times). \textbf{[Yes]} We provide results using boxplots and $avg\pm std$.
    \item A description of the computing infrastructure used. (e.g., type of GPUs, internal cluster, or cloud provider). \textbf{[Yes]} We provide this information in Appendix (Section Experimental Details).
  \end{enumerate}

  \item If you are using existing assets (e.g., code, data, models) or curating/releasing new assets, check if you include:
  \begin{enumerate}
    \item Citations of the creator If your work uses existing assets. \textbf{[Yes] }We cite all the datasets owners.
    \item The license information of the assets, if applicable. \textbf{[Not Applicable]}
    \item New assets either in the supplemental material or as a URL, if applicable. \textbf{[Not Applicable]}
    \item Information about consent from data providers/curators. \textbf{[Not Applicable]}
    \item Discussion of sensible content if applicable, e.g., personally identifiable information or offensive content. \textbf{[Not Applicable]}
  \end{enumerate}

  \item If you used crowdsourcing or conducted research with human subjects, check if you include:
  \begin{enumerate}
    \item The full text of instructions given to participants and screenshots. \textbf{[Not Applicable]}
    \item Descriptions of potential participant risks, with links to Institutional Review Board (IRB) approvals if applicable. \textbf{[Not Applicable]}
    \item The estimated hourly wage paid to participants and the total amount spent on participant compensation. \textbf{[Not Applicable]}
  \end{enumerate}

\end{enumerate}

\clearpage
\onecolumn
\appendix
\section{General Assumptions}
\label{appendix:general_assumptions}

Before presenting individual proofs and continuing further the discussion, we first outline the general assumptions that hold throughout this Appendix.  
These assumptions simplify notation while preserving full generality of the results and hold unless otherwise specified.

\subsection{Notation and Binning Assumptions.} We consider a multi-class classification setting, where $\mcX\subseteq\mathbb{R}^m$ is the feature space and $\mcY = \{0,\ldots, C-1\}$ is a finite target space with $C$ distinct labels. Let us assume we have access to a given dataset \( D = \{(\mbx_i, y_i)\}_{i=1}^n \) of input-output pairs drawn from an unknown joint distribution \( \mathcal{P} \) over \( \mathcal{X} \times \mathcal{Y} \). Each input \( \mbx_i \in \mcX \) is a feature vector of $m$ dimensions, and each label \( y_i \in \mcY \) has a corresponding one-hot encoded vector $\mathbf{y}_i$ indicating the correct class among the \( C \) possible classes. 
We consider a probabilistic classifier \( f \colon \mathcal{X} \rightarrow \Delta^C \), where \( \Delta^C \) is the \( (C - 1) \)-dimensional probability simplex. In words, a probabilistic classifier maps an input \( \mbx \) to a probability distribution over classes, i.e.,  \( f(\mbx) = \mathbf{\hat{p}} \in \Delta^C \), where each entry \( \mathbf{\hat{p}}_{k} = f_k(\mbx) \) of the predicted probability vector $\mathbf{\hat{p}}$ denotes the predicted probability of class \( k \).

Throughout this appendix, we present the theoretical results and corresponding proofs under a unified framework that  covers \textit{multi-dimensional binning}-based calibration metrics 
This formulation is adopted for clarity and compactness.
For completeness, at the end of each proof that requires it, we explicitly discuss how the same analytical results extend to the case of class-wise calibration metrics,
which involves only minor technical adjustments. 
Hence, the presented analysis extends naturally to both binning schemes.

\subsection{Conditioning Assumptions.}
\label{appendix:conditioning_ass}
All probabilistic bounds are derived under the following conditioning setup:
\begin{itemize}
    \item \textbf{Conditioning on features.}  
    We condition on the observed feature values \(\{\mbx_i\}_{i=1}^n\).  
    Since the binning rule \(\beta(\cdot)\) is deterministic, this also fixes the bin memberships \(\{I_b\}_{b=1}^{B}\).  
    Hence, after conditioning, the sets of indices per bin are deterministic.
    \item \textbf{Conditioning on kernel estimates.}  
    We condition on the kernel estimates \(\hat{\theta}_c(\mathbf{y}_i \mid \mbx_i)\), which are computed on an independent, arbitrarily large disjoint dataset.  
    For the evaluation set, these estimates depend only on the observed \(\mbx_i\); thus, once features are fixed, the estimates are deterministic as well. 
\end{itemize}

\subsection{Kernel Consistency Assumptions}
\label{appendix:kernel_assump}
In the following, we provide the underlying assumptions for the kernel estimator consistency:
\begin{enumerate}
    \item \textbf{Data Assumptions.} The evaluation set $D = \{(\mbx_i, y_i)\}_{i=1}^n$ is drawn independently and identically distributed (i.i.d.) from a joint distribution over a compact subset of $\mathcal{X} \times \mathcal{Y}$. The marginal probability density function $p(\mbx)$ and the true conditional probability function $p(\mathbf y|\mbx)$ are continuous and bounded on $\mathcal{X}$. 
    \item \textbf{Kernel Assumptions.} The kernel function $k$ is a non-negative, symmetric, and bounded function that integrates to one, i.e., $\int_{\mathbb{R}^d} k(u) \,du = 1$.
    \item \textbf{Bandwidth Assumptions.} The bandwidth parameter $\gamma_n$ is a positive sequence that depends on the sample size $n$ and satisfies the following conditions as $n \to \infty$:
    \begin{enumerate}
        \item $\gamma_n \to 0$ (the bandwidth shrinks).
        \item $n\gamma_n^d \to \infty$, where $d$ is the dimensionality of $\mathcal{X}$ 
    \end{enumerate}
\end{enumerate}
It is important to acknowledge that estimates do not constitute a point-wise unbiased approximation of the true conditional distribution.
Kernel estimators are known to suffer from both \textit{design bias}—a form of bias introduced by the distribution of the covariates $x$—and \textit{boundary bias}, which is particularly pronounced near the edges of the support or in low-density regions of the input space \citep{wasserman2006all}.
Nevertheless, NW estimators are known to be consistent in the mean squared error (MSE) sense under mild conditions on the kernel and bandwidth sequence \citep{DBLP:books/daglib/0035701, wasserman2006all}. Note that the MSE consistency is a \textit{stronger} notion that implies Mean Integrated Squared Error (MISE) and Mean Absolute Error (MAE) consistencies. As a consequence, the estimator asymptotically yields a statistically meaningful approximation of the target function:
\[
\mathbb{E}\!\left[\big(\hat{\theta}(\mathbf y\mid \mbx) - \Pr(\mathbf y\mid \mbx)\big)^2\right] \xrightarrow[n\to\infty]{} 0,
\]

\section{Proofs}
\label{app:proofs}
\subsection{Proof of Theorem 1}
\label{sec:proof_thm1}

\textbf{Statement.} Define the continuous Multidimensional Expected Calibration Error (MECE) as:
\[
M(f) = \sum_{y \in \mcY} p(y) \,\mathbb{E}_{\mbx \sim p(\mbx)} \left[ \left| \mathbb{E}[\mathbf{y} \mid \mbx]_y - f_y(\mbx) \right| \right], \notag
\]

If a model \( f \) satisfies local calibration, under standard regularity conditions there exists $k \in (0,1]$ such that  continuous $MECE$ is asymptotically upper bounded:
\[
M(f) \leq \varepsilon \cdot k
\]
\begin{proof}
We can rewrite this quantity by marginalizing over the input and output spaces:
\begin{align*}   
& \sum_{y \in \mcY} p(y) \,\mathbb{E}_{\mbx \sim p(\mbx)}\left[ \left| \mathbb{E}[\mathbf{y} \mid \mbx]_y - f_y(\mbx) \right| \right] 
= \int_{\mathcal{X}} \left[ \sum_{y \in \mathcal{Y}} p(y) \cdot \left| \mathbb{E}[\mathbf{y} \mid \mbx]_y - f_y(\mbx) \right| \right] p(\mbx) \, d\mbx \\
&\le \int_{\mathcal{X}} \left[ \sum_{y \in \mathcal{Y}} \max_{y'\in\mathcal{Y}}p(y') \cdot \left| \mathbb{E}[\mathbf{y} \mid \mbx]_y - f_y(\mbx) \right| \right] p(\mbx) \, d\mbx \\
\le &\max_{y\in\mathcal{Y}} \int_{\mathcal{X}} \left[ \sum_{y \in \mathcal{Y}} \cdot \left| \mathbb{E}[\mathbf{y} \mid \mbx]_y - f_y(\mbx) \right| \right] p(\mbx) \, d\mbx \notag
\end{align*}

By the triangle inequality we obtain:
\[
\sum_{y \in \mcY} p(y) \,\mathbb{E}_{x \sim p(x)}\left[ \left| \mathbb{E}[\mathbf{y} \mid \mbx]_y - f_y(\mbx) \right| \right] 
\leq \max_{y\in\mathcal{Y}}{{p(y)}} \int_{\mathcal{X}} \Bigg(\left\| \hat{\theta}(\mathbf y \mid \mbx) - f(\mbx) \right\|_1 + \left\| {\mathbb{E}}[\mathbf{y} \mid \mbx] - \hat{\theta}(\mathbf y \mid \mbx)  \right\|_1 \Bigg)\, p(x) \, dx. \notag
\]

Since \( f \) is locally calibrated, the first integrand term is bounded pointwise:
\[
\left\| \hat{\theta}(\mathbf y \mid \mbx) - f(\mbx) \right\|_1 \leq \varepsilon. \notag
\]
The second term, \( \int_{\mathcal{X}} \left\| \mathbb{E}[\mathbf{y} \mid \mbx] - \hat{\theta}(\mathbf y \mid \mbx) \right\|_1 p(\mbx) \, d\mbx \), vanishes asymptotically under mild regularity assumptions on the kernel and the data distribution (refer to Appendix~\ref{appendix:kernel_assump} for all the details):
\[
\lim_{n \to \infty} \int_{\mathcal{X}} \left\| \mathbb{E}[\mathbf{y} \mid \mbx] - \hat{\theta}(\mathbf y \mid \mbx) \right\|_1 p(x) \, dx = 0. \notag
\]

Hence, for sufficiently large \( n \), this second term can be made arbitrarily small. Denoting it by \( \delta_n \to 0 \), we provide the final bound for a constant $k \ge \max_{y\in\mathcal{Y}}{p(y)}$:
\[
\sum_{y \in \mcY} p(y) \, \mathbb{E}_{x \sim p(x)}\left[ \left| \mathbb{E}[\mathbf{y} \mid \mbx]_y - f_y(\mbx) \right| \right] \leq (\varepsilon + \delta_n)\cdot k. \notag
\]

As \( n \to \infty \), this yields:
\[
\sum_{y \in \mcY} p(y) \, \mathbb{E}_{x \sim p(x)}\left[ \left| \mathbb{E}[\mathbf{y} \mid \mbx]_y - f_y(x) \right| \right] \leq \varepsilon \cdot k. \qedhere
\]
\end{proof}
\subsection{Proof of Theorem 2}

\label{proof:thm2_proof}
\textbf{Statement.} Let $\mathcal{E}(D;\varphi;\beta)$ be a calibration metric satisfying \cref{def:metric_def}, evaluated on a sample $\{(\mbx_i,\mathbf y_i)\}_{i=1}^n$. 
Let $f$ be locally calibrated with error $\varepsilon$ relative to kernel estimates $\hat \theta(\mathbf y_i \mid \mbx_i)$ computed on an independent arbitrarily large auxiliary dataset. 
Conditioning on the evaluation inputs $\{\mbx_i\}_{i=1}^n$ and corresponding  consistent kernel estimates, for any $\delta \in (0,1)$, with probability at least $1-\delta$:

\[
\mathcal{E}(D;\varphi;\beta)
\le L_\varphi \sum_{b=1}^{m_B} \sum_{c=1}^C w_{b,c} \,\cdot\, \pi_c
\Bigg( \sqrt{\frac{\log(\frac{2Cm_B}{\delta})}{2 |\Psi(b,c)|}} \;+\; \varepsilon \Bigg) 
\]

where $\Psi(\cdot, \cdot)$ secelts a bin based on the index $b$ and possibly the label $c$.
\begin{proof}
We begin our proof by fixing a bin \(b\) and a class \(c\) and, given the per-instance kernel estimates \(\hat{\theta}_c(\mathbf{y}_i\mid \mbx_i)\), we define the local estimator average:
\[
\hat{\theta}_{b,c} \;:=\; \frac{1}{|B_b|}\sum_{i\in I_b} \hat{\theta}_c(\mathbf{y}_i\mid \mbx_i). \notag
\]
Then, for any fixed bin \(b\) and class \(c\),
\[
\varphi\big(\mathrm{freq}_{b,c},\mathrm{conf}_{b,c}\big)
= \varphi\big(\mathrm{freq}_{b,c},\hat{\theta}_{b,c}\big)
+ \big[\varphi(\mathrm{freq}_{b,c},\mathrm{conf}_{b,c})-\varphi(\mathrm{freq}_{b,c},\hat{\theta}_{b,c})\big]. \notag
\]
By the Lipschitz property,
\[
\big|\varphi(\mathrm{freq}_{b,c},\mathrm{conf}_{b,c})-\varphi(\mathrm{freq}_{b,c},\hat{\theta}_{b,c})\big|
\le L_\varphi\,\big|\mathrm{conf}_{b,c}-\hat{\theta}_{b,c}\big|. \notag
\]
Moreover, since \(\varphi\) is also Lipschitz in its first argument,
\[
\varphi(\mathrm{freq}_{b,c},\hat{\theta}_{b,c})
\le\big| \varphi(\mathrm{freq}_{b,c},\hat{\theta}_{b,c})-\varphi(\hat{\theta}_{b,c},\hat{\theta}_{b,c}) 
\big|
\le L_\varphi\,\big|\mathrm{freq}_{b,c}-\hat{\theta}_{b,c}\big|, \notag
\]
as a direct consequence of \(\varphi(t,t)=0\).

Combining the two we obtain:
\[
\varphi(\mathrm{freq}_{b,c},\mathrm{conf}_{b,c})
\le L_\varphi\Big( \big|\mathrm{freq}_{b,c}-\hat{\theta}_{b,c}\big|
+ \big|\mathrm{conf}_{b,c}-\hat{\theta}_{b,c}\big|\Big).
\]

Using (12) and (13), by pulling constants outside,
\[
\mathcal{E}(D;\varphi;\beta)
\le L_\varphi \sum_{b=1}^B w_b\sum_{c=1}^C \pi_c
\Big( \big|\mathrm{freq}_{b,c}-\hat{\theta}_{b,c}\big|\Big)
+  L_\varphi \sum_{b=1}^B w_b\sum_{c=1}^C \pi_c
\Big(\big|\mathrm{conf}_{b,c}-\hat{\theta}_{b,c}\big|\Big).
\]

We bound the two terms inside parentheses separately.

\paragraph{(ii) Miscalibration: \(|\mathrm{conf}_{b,c}-\hat{\theta}_{b,c}|\).}
\[
\mathrm{conf}_{b,c}-\hat{\theta}_{b,c}
= \frac{1}{|B_b|}\sum_{i\in I_b}\big(\mathbf{\hat{p}}_{i,c}-\hat{\theta}_c(\mathbf y_i \mid \mbx_i)\big). \notag
\]
By the local calibration assumption, for every instance \(i\) we have
\(\| \mathbf{\hat{p}}_i - \hat{\theta}(\mathbf{y}_i\mid \mbx_i)\|_1 \le \varepsilon\). Consequently each coordinate satisfies
\(|\mathbf{p}_{i,c}-\hat{\theta}_c(\mathbf{y}_i\mid \mbx_i)| \le \varepsilon\), hence
\[
\big|\mathrm{conf}_{b,c}-\hat{\theta}_{b,c}\big|
\le \frac{1}{|B_b|}\sum_{i\in I_b} |\mathbf{\hat{p}}_{i,c}-\hat{\theta}_c(\mathbf{y}_i\mid \mbx_i)|
\le \varepsilon. 
\]

\paragraph{(i) Empirical fluctuation: \(|\mathrm{freq}_{b,c}-\hat{\theta}_{b,c}|\).}
For fixed \(b,c\),
\[
\mathrm{freq}_{b,c}-\hat{\theta}_{b,c}
= \frac{1}{|B_b|}\sum_{i\in I_b}\big(\mathbf{1}\{y_i=c\}-\hat{\theta}_c(\mathbf{y}_i\mid \mbx_i)\big). \notag
\]
In the following we apply Hoeffding inequality to bound the \textbf{Empirical fluctuation} term. More precisely, to apply Hoeffding, we need independent, bounded, zero-mean summands. 

Under the \textbf{conditioning assumptions} (Appendix~\ref{appendix:conditioning_ass}), the only source of randomness in the term
\[
\mathrm{freq}_{b,c}-\hat{\theta}_{b,c}
= \frac{1}{|B_b|}\sum_{i\in I_b}\big(\mathbf{1}\{y_i=c\}-\hat{\theta}_c(\mathbf{y}_i\mid \mbx_i)\big) \notag
\]
is the label variables $\{y_i\}_{i \in I_b}$. For each $i$, the summand satisfies:  

\begin{enumerate}
    \item \textbf{Zero mean:}
    \[
    \mathbb{E}\!\left[\mathbf{1}\{y_i=c\} - \hat{\theta}_c(\mathbf{y}_i\mid \mbx_i) \;\middle|\; \mbx_i\right] 
    = P(y_i=c \mid \mbx_i) - \hat{\theta}_c(\mathbf{y}_i\mid \mbx_i) \approx 0, \notag
    \]
    and exactly zero if the estimator is consistent in mean absolute error (MAE) (refer to Appendix~\ref{appendix:kernel_assump} for all the details).  
    \item \textbf{Independence:} the pairs $(\mbx_i,y_i)$ are i.i.d., and conditioning on the $\mbx_i$ leaves the labels $\{y_i\}$ independent.  
    \item \textbf{Boundedness:} both $\mathbf{1}\{y_i=c\}$ and $\hat{\theta}_c(\mathbf{y}_i\mid \mbx_i)$ lie in $[0,1]$, hence their difference lies in $[-1,1]$.  
\end{enumerate}

Hoeffding’s inequality applies to their average, yielding the desired concentration bound.  
More precisely, Hoeffding's inequality gives, for any \(\tau>0\),
\[
\Pr\Big(\Big|\mathrm{freq}_{b,c}-\hat{\theta}_{b,c}\Big|>\tau\Big) \le 2\exp(-2 |B_b| \tau^2). \notag
\]
Choosing \(\tau_b=\sqrt{\tfrac{\log(\frac{2Cm_B}{\delta})}{2 |B_b|}}\) and applying the union bound over the \(m_B\cdot C\) bin–class pairs yields: with probability at least \(1-\delta\),
\[
\forall \; b,c:\qquad \big|\mathrm{freq}_{b,c}-\hat{\theta}_{b,c}\big|
\le \sqrt{\frac{\log(\frac{2Cm_B}{\delta})}{2 |B_b|}}.
\] 
Now insert (15) and (16) into (14). With probability at least \(1-\delta\),
\begin{equation}
\label{eqn:thm2_bound}
\boxed{%
\mathcal{E}(D;\varphi;\beta)
\le L_\varphi \sum_{b=1}^{m_B} \sum_{c=1}^C w_b \cdot\pi_c
\Bigg( \sqrt{\frac{\log(\frac{2Cm_B}{\delta})}{2 |B_b|}} + \varepsilon \Bigg)} 
\end{equation} 
Therefore achieving the form of \cref{thm:generic_bound} when \(\Psi(b,c) = B_{b}\) and $w_{b,c}=w_b$.
%
%
%
%
%
%
\paragraph{Class-wise metrics.}
In the context of class-wise metrics, 
the only difference is the introduction of a dependence between bins' cardinalities and the classes: $|B_b| \to |B_{b,c}|$. Then, a bound of same style applies and with probability at least $1-\delta$:
\begin{align}
\mathcal{E}(D;\varphi;\beta)
&\le L_\varphi \sum_{b=1}^{m_B} \sum_{c=1}^C w_{b,c} \cdot \pi_c
\Bigg( \sqrt{\frac{\log(\frac{2Cm_B}{\delta})}{2 |B_{b,c}|}} + \varepsilon \Bigg) 
\notag
\end{align}
and by setting \(\Psi(b,c) = B_{b,c}\)  we obtain a bound of the same form of \cref{thm:generic_bound}. \end{proof}
\paragraph{Discussion on finite-sample kernel estimates.}
The proof above assumes that the reference kernel estimates are computed on an arbitrarily large disjoint set, so that they may be treated as deterministic. When the Nadaraya–Watson estimator is instead computed on a finite sample, additional stochastic error arises.

Fix a bin $b$, class $c$, and anchor point $\mbx_i$. The kernel estimator can be written as
\[
\hat\theta_{i,c}
=
\sum_{j\in I_b} w_{ij}\,\mathbf 1\{y_j=c\},
\qquad
w_{ij}
=
\frac{k_{\gamma_n}(\mbx_i,\mbx_j)}
{\sum_{\ell\in I_b} k_{\gamma_n}(\mbx_i,\mbx_\ell)} .
\]

Conditioning on the inputs $\{\mbx_j\}_{j\in I_b}$, the weights $w_{ij}$ are deterministic and the only randomness stems from the labels $\{y_j\}_{j\in I_b}$. Since the label indicators $\mathbf 1\{y_j=c\}$ are independent and bounded in $[0,1]$, a weighted Hoeffding inequality applies. In particular,
\[
\Pr\!\left(
\big|\hat\theta_{i,c}-\mathbb E[\hat\theta_{i,c}\mid X]\big|
\ge \tau
\;\middle|\; X
\right)
\le
2\exp\!\left(-2\tau^2\, n_i^{\mathrm{eff}}\right),
\]
where the effective sample size is
\[
n_i^{\mathrm{eff}}
:=
\frac{1}{\sum_{j\in I_b} w_{ij}^2}.
\]

Applying a union bound over all $m_B C$ bin–class pairs yields that, with probability at least $1-\delta$,
\[
\forall b,c:\quad
\big|\hat\theta_{b,c}-\mathbb E[\hat\theta_{b,c}\mid X]\big|
\le
\sqrt{\frac{\log(\frac{2Cm_B}{\delta})}{2\, n_{b}^{\mathrm{eff}}}},
\]
where $n_b^{\mathrm{eff}}$ denotes a lower bound on the effective sample size within bin $b$.

Now decompose
\[
\big|\mathrm{freq}_{b,c}-\hat\theta_{b,c}\big|
=
\big|\mathrm{freq}_{b,c}- \mathbb E[\hat\theta_{b,c}\mid X]
-\hat\theta_{b,c}
+ \mathbb E[\hat\theta_{b,c}\mid X]\big|
\]
and apply the triangle inequality:
\[
\big|\mathrm{freq}_{b,c}-\hat\theta_{b,c}\big| \le
\big|\mathrm{freq}_{b,c}- \mathbb E[\hat\theta_{b,c}\mid X]\big|
+
\big| \hat\theta_{b,c}- \mathbb E[\hat\theta_{b,c}\mid X]\big|.
\]

The first term is the standard empirical fluctuation term and satisfies
\[
\big|\mathrm{freq}_{b,c}- \mathbb E[\hat\theta_{b,c}\mid X]\big|
\le
\sqrt{\frac{\log(\frac{2Cm_B}{\delta})}{2\, |B_b|}}.
\]

The second term satisfies
\[
\big| \hat\theta_{b,c}- \mathbb E[\hat\theta_{b,c}\mid X]\big|
\le
\sqrt{\frac{\log(\frac{2Cm_B}{\delta})}{2\, n_b^{\mathrm{eff}}}}.
\]

Since $n_b^{\mathrm{eff}} \le |B_b|$, the two terms are of the same order and we may upper bound their sum by
\[
\big|\mathrm{freq}_{b,c}-\hat\theta_{b,c}\big|
\le
2 \;
\sqrt{\frac{\log(\frac{2Cm_B}{\delta})}{2\, n_b^{\mathrm{eff}}}}.
\]

Moreover, under standard smoothness assumptions on the conditional class probabilities (Appendix~\ref{appendix:kernel_assump}), the kernel-smoothed conditional probability differs from the true conditional probability by a bias of order $\mathcal O(\gamma_n^2)$ \citep{wasserman2006all}. Combining stochastic concentration and smoothing bias yields
\[
\big|\mathrm{freq}_{b,c}-\hat\theta_{b,c}\big|
\le
2 \;
\sqrt{\frac{\log(\frac{2Cm_B}{\delta})}{2\, n_b^{\mathrm{eff}}}}
+
\mathcal O(\gamma_n^2).
\]

Therefore the bound in \cref{eqn:thm2_bound} becomes:
\[
\Pr\Bigg(
\mathcal{E}(D;\varphi;\beta)
\le
L_\varphi
\sum_{b=1}^{m_B}\sum_{c=1}^C w_b \cdot \pi_c \Big(\varepsilon +2 \;
\sqrt{\frac{\log(\frac{2Cm_B}{\delta})}{2\, n_b^{\mathrm{eff}}}} 
+
\mathcal O(\gamma_n^2)
\Big)\Bigg)
\ge
1-\delta.
\]

The additional $\mathcal O(\gamma_n^2)$ term vanishes as $n\to\infty$ under the kernel consistency assumptions.
\subsection{Proof of Corollary 1.} 
\label{subsec:proof_cor1}

\textbf{Statement.} If a classifier $f$ satisfies $\rho$-\textit{perfect local calibration}, the error \( \mathcal{E}(D;\varphi;\beta) \) becomes purely stochastic fluctuation. For any \(\delta \in [0,1]\), with probability at least $1-\delta$ it holds that: 
\[
\limsup_{\rho \to 0} \mathcal{E}(D;\varphi;\beta) \leq L_\varphi\sum_{b=1}^{m_B} \sum_{c=1}^C w_{b,c}\,\cdot\,\pi_c\, \, \sqrt{ \frac{ \log(\frac{2Cm_B}{\delta}) }{ 2 |\Psi(b,c)| } }
\]
\begin{proof}
Recall that by the local calibration assumption, for every instance \( i \in D \) we have
\(\| \mathbf{\hat{p}}_i - \hat{\theta}(\mathbf{y}_i\mid \mbx_i)\|_1 \le \varepsilon\). Consequently, each coordinate satisfies
\(|\mathbf{\hat{p}}_{i,c}-\hat{\theta}_c(\mathbf{y}_i\mid \mbx_i)| \le \varepsilon\), hence:
\[
\big|\mathrm{conf}_{b,c}-\hat{\theta}_{b,c}\big|
\le \frac{1}{|B_{b,c}|}\sum_{i\in I_{b,c}} |\mathbf{\hat{p}}_{i,c}-\hat{\theta}_c(\mathbf{y}_i\mid \mbx_i)|
\le \varepsilon. \notag
\]

If additionally the model \( f \) satisfies \(\rho\)-perfect uniform local calibration. Then, for every instance \( i \in D \) and class \( c \in \{1, \ldots, C\} \), the absolute calibration error is bounded:
\[
\left| \mathbf{\hat{p}}_{i,c} -  \hat{\theta}_c(\mathbf{y}_i\mid \mbx_i) \right| \leq L\cdot\rho\notag
\]
where \( L\cdot\rho \) is the maximum variation in the predicted probability for class \( c \) within the isotropic neighborhood of radius $\rho$. In this context the miscalibration error can be further reduced: 
\[
 \left\| \hat{\theta}(\mathbf{y}_i\mid \mbx_i) - \mathbf{\hat{p}}_i \right\|_1 \leq C \cdot L\cdot\rho \notag 
\]
Substituting back into (6) we obtain:

\[
\Pr\Bigg(\mathcal{E}(D;\varphi;\beta)
\le L_\varphi\sum_{b=1}^{m_B}\sum_{c=1}^C w_{b,c} \cdot \pi_c\Bigg(\sqrt{\frac{\log(\frac{2Cm_B}{\delta})}{2 |\Psi(b,c)|}}+ C\cdot L\cdot\rho\Bigg)\Bigg)\ge 1-\delta.\notag
\]


\textbf{Conclusion:}  In the limit of $\rho$-perfect local calibration, the calibration error reduces to pure stochastic fluctuation: 
\[
\boxed{%
\limsup_{\rho \to 0} \Pr\Bigg(\mathcal{E}(D;\varphi;\beta) \leq L_\varphi\sum_{b=1}^{m_B} \sum_{c=1}^C w_{b,c}\cdot\pi_c \, \sqrt{ \frac{ \log(\frac{2Cm_B}{\delta}) }{ 2 |\Psi(b,c)| } }\Bigg) \geq 1- \delta}. \tag{7} 
\]
\end{proof}

\subsection{Proof of Theorem 3} 
\textbf{Statement.} Let \( f\) be composed of a feature extractor \( \phi\) and classification layer \( g\).
Let $f$ be locally calibrated with error $\varepsilon$ relative to kernel estimates $\hat \theta(\mathbf y_i \mid \mbx_i)$ computed on an independent arbitrarily large auxiliary dataset. 
Let the LCE be computed on $\phi(\mbx_i)$. 
Then, conditioning on the evaluation inputs $\{\mbx_i\}_{i=1}^n$ and on the corresponding consistent kernel estimates, for any \(\delta \in [0,1]\), with at least probability  \(1-\delta\): 

\begin{align*}
\mathrm{LCE} \le \; \varepsilon \;+\; \frac{2}{n}\sum_{b=1}^{m_B}\sum_{i \in I_b}\sqrt{\frac{\log\!\bigl(\frac{2nC}{\delta}\bigr)}{2n_i^{\mathrm{eff}}}} \: + \:
\frac{L}{n}\sum_{b=1}^{m_B}
\mathbb{E}\!\Big[\sum_{i\in I_b}\sum_{j\in I_b} w_{i,j}\|\phi(\mbx_j)-\phi(\mbx_i)\|_1\Big]s
\end{align*}
where \(
n_i^{\mathrm{eff}}= \frac{1}{\sum_{j \in I_b} w_{i,j}^2}
\), with $w_{i,j}=\frac{k_{\gamma}(\phi(\mbx_i), \phi(\mbx_j))}{\sum_j k_{\gamma}(\phi(\mbx_i),\phi(\mbx_j))}$.

\label{sec:proof_thm3}
\begin{proof}
Let $k_\gamma(\mbx_i,\mbx_j)$ kernel functions to obtain the kernel-weighted mean of both the empirical frequencies and the predicted probabilities for given anchor point $\mbx_i$:
\begin{align*}
&\hat{\theta}(\mathbf{y}_i\mid \mbx_i) := \sum_{j \in I_b} \frac{k_\gamma(\mbx_i,\mbx_j)}{\sum_{j \in I_b} k_\gamma(\mbx_i,\mbx_j)} \, \mathbf{y}_{j}, \\
    &\hat{\theta}(\mathbf{\hat{p}}_i\mid \mbx_i) := \sum_{j \in I_b} \frac{k_\gamma(\mbx_i,\mbx_j)}{\sum_{j \in I_b} k_\gamma(\mbx_i,\mbx_j)} \, \mathbf{\hat{p}}_{j}.
\end{align*}

For a given bin $b$ we write the value of LCE:
\[ 
\text{LCE} = \frac1{C}\sum_{b=1}^{m_B} \frac{|B_b|}{n} \frac1{|B_b|}\sum_{i\in b}\left\|\frac{\sum_{j\in b} \bigl(\mathbf{\hat{p}}_{j} - \mathbf{y}_j \bigr)\, k_\gamma(\mbx_i, \mbx_j)}
{\sum_{j \in b} k_\gamma(\mbx_i, \mbx_j)}\right\|_1
= \frac1{C}\sum_{b=1}^{m_B} \frac{|B_b|}{n} \frac1{|B_b|}\sum_{i\in b}\|\hat{\theta}(\mathbf{\hat{p}}_i\mid \mbx_i) - \hat{\theta}(\mathbf{y}_i\mid \mbx_i) \|_1 \notag
\]

In the following, we bound the deviation coordinate-wise and obtain a uniform statement over all classes $c\in[C]$.
Fix a bin $b$ and consider one coordinate $c$.
By triangle inequality,
\begin{align*}
\frac{1}{|B_b|}\sum_{i\in I_b}
\big|\hat\theta(\hat p_{i,c}\mid \mbx_i)-\hat\theta(y_{i,c}\mid \mbx_i)\big|
&\le
\frac{1}{|B_b|}\sum_{i\in I_b}\big|\hat\theta(\hat p_{i,c}\mid \mbx_i)-\hat p_{i,c}\big|
+
\frac{1}{|B_b|}\sum_{i\in I_b}\big|\hat p_{i,c}-\hat\theta(y_{i,c}\mid \mbx_i)\big|
\end{align*}

By local calibration with respect to the (infinite-sample) kernel target $\theta(\cdot\mid \mbx_i)$,
$\|\hat{\mathbf p}_i-\theta(\mathbf y\mid \mbx_i)\|_1\le \varepsilon$,
hence for every $c$,
$|\hat p_{i,c}-\theta( y_{i,c}\mid \mbx_i)|\le \varepsilon$.
Therefore, for each $i\in I_b$,
\[
\big|\hat\theta(y_{i,c}\mid \mbx_i)-\hat p_{i,c}\big|
\le
\big|\hat\theta(y_{i,c}\mid \mbx_i)-\theta(y_{i,c}\mid \mbx_i)\big|
+\varepsilon
\]

Moreover, conditioning on the inputs, $\hat\theta(y_{i,c}\mid \mbx_i)$ is a weighted average of bounded independent label indicators.
Thus, we apply Hoeffding's inequality and for each fixed $(i,c)$ and with probability at least $1-\delta'$,
\[
\big|\hat\theta(y_{i,c}\mid \mbx_i)-\theta(y_{i,c}\mid \mbx_i)\big|
\le
\sqrt{\frac{\log(\frac{2}{\delta'})}{2\,n_i^{\mathrm{eff}}}},
\qquad
n_i^{\mathrm{eff}}:=\Big(\sum_{j\in I_b}w_{ij}^2\Big)^{-1}
\]
Setting $\delta'=\delta/(nC)$ and applying a union bound over all anchors $i$ and classes $c$
implies that with probability at least $1-\delta$,
\[
\forall i,c:\quad
\big|\hat\theta(y_{i,c}\mid \mbx_i)-\theta(y_{i,c}\mid \mbx_i)\big|
\le
\sqrt{\frac{\log(\frac{2nC}{\delta})}{2\,n_i^{\mathrm{eff}}}}
\]

Averaging over $i\in I_b$ yields
\[
\frac{1}{|B_b|}\sum_{i\in I_b}\big|\hat\theta(y_{i,c}\mid \mbx_i)-\hat p_{i,c}\big|
\le
\varepsilon
+
\frac{1}{|B_b|}\sum_{i\in I_b}\sqrt{\frac{\log(\frac{2nC}{\delta})}{2n_i^{\mathrm{eff}}}}
\]

Before proceeding let us rewrite: 
\begin{align*}
& \big|\,\hat{\theta}({\hat{p}}_{i,c}\mid \mbx_i)-{\hat{p}}_{i,c}\,\big| = \Big|\,\sum_{j\in I_b} w_{ij}{\hat{p}}_{j,c} - {\hat{p}}_{i,c}\Big|\, =  \big|\,\sum_{j\in I_b} w_{ij}({\hat{p}}_{j,c} - {\hat{p}}_{i,c}) + \sum_{j\in I_b} w_{ij}{\hat{p}}_{i,c} - {\hat{p}}_{i,c} \,\big| = \\
& \big|\,\sum_{j\in I_b} w_{ij}({\hat{p}}_{j,c} - {\hat{p}}_{i,c}) + (\underbrace{\sum_{j\in I_b} w_{ij}}_{=1} - 1){\hat{p}}_{i,c} \,\big|  = \big|\,\sum_{j\in I_b} w_{ij}({\hat{p}}_{j,c} - {\hat{p}}_{i,c})\,\big| 
\notag
\end{align*}
In addition note that:
\[
 \frac{1}{|B_b|}\sum_{i \in I_b} \big|\,\sum_{j\in I_b} w_{ij}({\hat{p}}_{j,c} - {\hat{p}}_{i,c})\,\big|  \leq \frac{1}{|B_b|}\sum_{i \in I_b} \underbrace{\sum_{j \in I_b} w_{i,j}\big|\,{\hat{p}}_{j,c} - {\hat{p}}_{i,c}\,\big|}_{Z_i}  \notag
\] 

Again we follow the \textbf{conditioning assumptions} of Appendix~\ref{appendix:conditioning_ass}.
Please note that each coordinate-wise distance satisfies \({\hat{p}}_{j,c}-{\hat{p}}_{i,c}\in[-1,1]\), hence \(Z_i\in[0,1]\). We can now apply the weighted version of Hoeffding’s inequality to each centered quantity $Z_i-\mathbb{E}[Z_i]$, conditioning on the kernel weights and features to obtain zero-mean summands. The union bound then yields a simultaneous statement over anchors.
We proceed fixing a bin $b$ with index set $I_b$ of size $|B_b|$. For each anchor $i \in I_b$ define the \emph{effective sample size} associated with the weights:
\[
n_i^{\mathrm{eff}} \;:=\; \frac1{\sum_{j \in I_b} w_{i,j}^2}
\notag
\]
Applying weighted Hoeffding’s to $Z_i-\mathbb{E}[Z_i]$ for zero mean Hoeffding assumption: 

\[
\Pr\Bigg( Z_i-\mathbb{E}[Z_i] \geq \tau \Bigg) \le
\text{exp}\Bigg(-2n_i^{\text{eff}}\tau^2 \Bigg) \notag
\]

And, by setting $\delta' = \delta/nC$ we have:

\[
\Pr\Bigg( Z_i-\mathbb{E}[Z_i] \le \sqrt{\frac{\log\!\bigl(\frac{1}{\delta'}\bigr)}{2n_i^{\mathrm{eff}}}}\Bigg) \geq 1-\delta' \notag
\]
The union bound over all $n$ anchors gives that with probability at least $1-\delta$ for every anchor:

\[Z_i - \mathbb{E}[Z_i] \le \sqrt{\frac{\log\!\bigl(\frac{nC}{\delta}\bigr)}{2n_i^{\mathrm{eff}}}} \notag
\]
Averaging for all anchors in the bin:

\[
\Pr\Bigg(\frac1{|B_b|}\sum_{i \in I_b} Z_i \le \frac1{|B_b|} \sum_{i \in I_b}\mathbb{E}[Z_i] + \frac1{|B_b|}\sum_{i \in I_b} \sqrt{\frac{\log\!\bigl(\frac{nC}{\delta}\bigr)}{2n_i^{\mathrm{eff}}}}\Bigg) \geq 1-\delta \notag
\]

We conclude the proof by providing a  bound for the expectation in the context of a neural network classifier $\phi(\cdot)$. 
More precisely, by the Lipschitz continuity of the softmax~\citep{DBLP:journals/corr/abs-1704-00805},
\[
\frac{1}{|B_b|}\sum_{i \in I_b} \sum_{j \in I_b} w_{i,j}\,\big|\,{\hat{p}}_{j,c} - {\hat{p}}_{i,c}\,\big| \leq \frac{L}{|B_b|}\sum_{i \in I_b} \sum_{j \in I_b} w_{i,j}\|\phi(\mbx_j) - \phi(\mbx_i) \|_1 \notag
\]
More precisely, $L\leq 1$ if the kernel estimates are obtain using logits as inputs. If instead the kernel is applied to $\phi(\cdot)$ mapped to logits via $z=Wh+b$ then $L\leq \max_{1 \leq j \leq n}\sum_{i=1}^m |W_{ij}|$. We now define the kernel-weighted local radius or \(\phi(\mbx_i)\):
\[
R_i := \sum_{j \in I_b} w_{i,j}\|\phi(\mbx_j) - \phi(\mbx_i) \|_1.\notag
\]
Thus \(Z_i \le L R_i\). Taking expectation over the sampling of points in the bin: 

\[
\mathbb{E}[Z_i] \le L\,\mathbb{E}[R_i]. \notag
\]
 Combining we obtain that with probability at least \(1-\delta\),

\[
\frac{1}{|B_b|}\sum_{i\in I_b}\big|\,\hat{\theta}({\hat{p}}_{i,c} \mid \mbx_i)-{\hat{p}}_{i,c}\,\big|
\le \frac{L}{|B_b|}\,
\mathbb{E}\!\Big[\sum_{i\in I_b}\sum_{j \in I_b} w_{i,j}\|\phi(\mbx_j) - \phi(\mbx_i) \|_1.\Big]
+\frac1{|B_b|}\sum_{i\in I_b}\sqrt{\frac{\log\!\bigl(\tfrac{nC}{\delta}\bigr)}{2n_i^{\mathrm{eff}}}} \notag
\]
\textbf{Conclusion:} as \(\sum_{c=1}^C 1/C = 1\), averaging over bins (weight \(|B_b|/n\)) yields the final bound: 
\[
\boxed{%
\Pr\Bigg(\mathrm{LCE} \le \; \varepsilon \;+\;  
\frac{L}{n}\sum_{b=1}^{m_B}
\mathbb{E}\!\Big[\sum_{i\in I_b}\sum_{j\in I_b} w_{i,j}\|\phi(\mbx_j)-\phi(\mbx_i)\|_1\Big]
\;+\; \frac{2}{n}\sum_{b=1}^{m_B}\sum_{i \in I_b}\sqrt{\frac{\log\!\bigl(\tfrac{2nC}{\delta}\bigr)}{2n_i^{\mathrm{eff}}}}\Bigg) \ge 1-\delta
}
\] 
\end{proof}
The bound decomposes into a \emph{bias term} $\mathbb{E}[Z]$, which depends on the kernel radius through the weights $w_{i,j}$, and an average \emph{variance term} that scales as $1/\sqrt{2n_i^{\mathrm{eff}}}$. 
Smaller kernel radii yield more concentrated weights: this reduces bias but also decreases $n_i^{\mathrm{eff}}$, thereby inflating the average variance. 
Conversely, larger kernels spread the weights more evenly, which decreases variance at the expense of bias. 
This captures the bias–variance tradeoff.
\paragraph{Class-wise metric.}
In the context of class-wise $LCE$ the only difference is the introduction of a dependence between bins' cardinalities and the classes. Then, with probability at least $1-\delta$, the bound applies with the same exact form under the following minor changes:
\[
\mathrm{cwLCE}\;\le\;
\varepsilon
\;+\;
\frac{L}{nC}\sum_{c=1}^C\sum_{b=1}^{m_B}
\mathbb{E}\!\Bigg[\sum_{i\in I_{b,c}}\sum_{j\in I_{b,c}} w_{i,j}\,\|\phi(\mbx_j)-\phi(\mbx_i)\|_1\Bigg]
\;+\;
\frac{2}{nC}\sum_{c=1}^C\sum_{b=1}^{m_B}\sum_{i\in I_{b,c}}
\sqrt{\frac{\log\!\bigl(\tfrac{2\,n\,C}{\delta}\bigr)}{2\,n^{\mathrm{eff}}_{i,c}}}
\]
\subsection{Proof of Theorem 4}
\label{sec:proof_thm4}

\textbf{Statement.} Let $f$ be a probabilistic classifier and let $\hat{\theta}(\cdot \mid \mathbf{x})$ be a kernel estimator of $p(\cdot \mid \mathbf{x})$ that is consistent in mean squared error. Then, under standard regularity conditions:
\begin{align}
    \lim_{n \to \infty} \frac{1}{n} \sum_{i=1}^n \text{d}_{\mathrm{JSD}} \left( \mathbf{\hat{p}}_i, \hat{\theta}(\mathbf y\mid \mbx_i \right) = 
    \lim_{n \to \infty} \frac{1}{n} \sum_{i=1}^n \text{d}_{\mathrm{JSD}} \left( \mathbf{\hat{p}}_i,\mathbf{p}_i\right)
\end{align}

\begin{proof}
Let \( \hat{P}_i \in \Delta^{C} \) be the softmax prediction for input \( \mbx_i \), and let \( \hat{Q}_i \in \Delta^{C} \) be a consistent estimator in the mean integrated squared error sense (MISE) (refer to Appendix ~\ref{appendix:kernel_assump} for a detailed description of the underlying assumptions) of the true conditional distribution \( Q_i = \Pr(\mathbf y_i \mid \mbx_i) \), meaning:
\[
\lim_{n \to \infty} \frac{1}{n} \sum_{i=1}^n \|\hat{Q}_i - Q_i\|_1 = 0. \notag
\]
Then, it suffices to show that the average Jensen-Shannon distance computed using \( \hat{Q}_i \) converges to the one computed using the true distribution:
\[
\lim_{n \to \infty} \frac{1}{n} \sum_{i=1}^n \text{d}_{\mathrm{JSD}}(\hat{P}_i \| \hat{Q}_i) = \lim_{n \to \infty} \frac{1}{n} \sum_{i=1}^n \text{d}_{\mathrm{JSD}}(\hat{P}_i \| Q_i),
\]
where \( \text{d}_{\mathrm{JSD}}(P \| Q) := \sqrt{\mathrm{JSD}(P \| Q)} \) denotes the Jensen-Shannon distance.

Since the Jensen-Shannon distance \( \text{d}_{\mathrm{JSD}} \) is a metric, it satisfies the triangle inequality:
\[
\text{d}_{\mathrm{JSD}}(\hat{P}_i \| \hat{Q}_i) \leq \text{d}_{\mathrm{JSD}}(\hat{P}_i \| Q_i) + \text{d}_{\mathrm{JSD}}(Q_i \| \hat{Q}_i). \notag
\]
Averaging over \( i \), we obtain:
\[
\frac{1}{n} \sum_{i=1}^n \text{d}_{\mathrm{JSD}}(\hat{P}_i \| \hat{Q}_i) \leq \frac{1}{n} \sum_{i=1}^n \text{d}_{\mathrm{JSD}}(\hat{P}_i \| Q_i) + \frac{1}{n} \sum_{i=1}^n \text{d}_{\mathrm{JSD}}(Q_i \| \hat{Q}_i). \notag
\]

We now apply an inequality that relates the Jensen-Shannon divergence to the total variation distance. For any pair of categorical distributions \( Q, \hat{Q} \), it holds that:
\[
\mathrm{JSD}(Q \| \hat{Q}) \leq \frac{\text{log}_b(2)}{2} \| Q - \hat{Q} \|_1. \notag
\]
which depends on the $\text{log}$ basis $b$ used to compute $\text{JSD}$.
Taking square roots and averaging, and using Jensen’s inequality for the concave square root function:
\[
\frac{1}{n} \sum_{i=1}^n \text{d}_{\mathrm{JSD}}(Q_i \| \hat{Q}_i) = \frac{1}{n} \sum_{i=1}^n \sqrt{\mathrm{JSD}(Q_i \| \hat{Q}_i)} \leq \sqrt{ \frac{1}{n} \sum_{i=1}^n \mathrm{JSD}(Q_i \| \hat{Q}_i) } \leq \sqrt{ \frac{\text{log}_b(2)}{2n} \sum_{i=1}^n \| Q_i - \hat{Q}_i \|_1 }. \notag
\]
By the consistency assumption of kernel estimator,
\[
\frac{1}{n} \sum_{i=1}^n \| Q_i - \hat{Q}_i \|_1 \to 0 \quad \text{as } n \to \infty, \notag
\]
and therefore,
\[
\frac{1}{n} \sum_{i=1}^n \text{d}_{\mathrm{JSD}}(Q_i \| \hat{Q}_i) \to 0. \notag
\]
Combining we obtain:
\[
\limsup_{n \to \infty} \frac{1}{n} \sum_{i=1}^n \text{d}_{\mathrm{JSD}}(\hat{P}_i \| \hat{Q}_i) \leq \lim_{n \to \infty} \frac{1}{n} \sum_{i=1}^n \text{d}_{\mathrm{JSD}}(\hat{P}_i \| Q_i). \notag
\]

We now prove the reverse inequality. Again, using the triangle inequality:
\[
\text{d}_{\mathrm{JSD}}(\hat{P}_i \| Q_i) \leq \text{d}_{\mathrm{JSD}}(\hat{P}_i \| \hat{Q}_i) + \text{d}_{\mathrm{JSD}}(\hat{Q}_i \| Q_i), \notag
\]
and therefore:
\[
\frac{1}{n} \sum_{i=1}^n \text{d}_{\mathrm{JSD}}(\hat{P}_i \| Q_i) \leq \frac{1}{n} \sum_{i=1}^n \text{d}_{\mathrm{JSD}}(\hat{P}_i \| \hat{Q}_i) + \frac{1}{n} \sum_{i=1}^n \text{d}_{\mathrm{JSD}}(\hat{Q}_i \| Q_i). \notag
\]
As before, by symmetry:
\[
\frac{1}{n} \sum_{i=1}^n \text{d}_{\mathrm{JSD}}(\hat{Q}_i \| Q_i) \to 0. \notag
\]
Combining we obtain:
\[
\liminf_{n \to \infty} \frac{1}{n} \sum_{i=1}^n \text{d}_{\mathrm{JSD}}(\hat{P}_i \| \hat{Q}_i) \geq \lim_{n \to \infty} \frac{1}{n} \sum_{i=1}^n \text{d}_{\mathrm{JSD}}(\hat{P}_i \| Q_i). \notag
\]

Which concludes our proof:
\[
\boxed{\lim_{n \to \infty} \frac{1}{n} \sum_{i=1}^n \text{d}_{\mathrm{JSD}}(\hat{P}_i \| \hat{Q}_i) = \lim_{n \to \infty} \frac{1}{n} \sum_{i=1}^n \text{d}_{\mathrm{JSD}}(\hat{P}_i \| Q_i)}. \qedhere
\]
\end{proof}
%
%
%

%
%
%
%

\section{Further Discussion}
\label{appendix:further_discussion}

\subsection{Extension of Theorem~\ref{thm:generic_bound} to ECCE}
\label{sec:extension_ecce}
The subsequent analysis aims to extend the applicability of Theorem~\ref{thm:generic_bound} to the specific class of cumulative binning-based metrics, with a focus on $ECCE$ (Expected Cumulative Calibration Error).
Unlike standard binning metrics, which directly compare per-bin statistics (as defined in \cref{def:metric_def}), cumulative binning metrics operate on the cumulative sums of per-bin statistics.
Despite this systematic difference, we demonstrate that cumulative binning metrics, specifically $ECCE$, admit an upper bound of an analogous form to that presented in Theorem~\ref{thm:generic_bound}, with slight discrepancies due to the different procedure of iteration over bins being applied. This establishes cumulative binning metrics as a special case under the unifying bound structure provided.

Let us consider a multi-class classification setting with label space \( \mathcal{Y} = \{1, \ldots, C\} \), and assume that the dataset \( D = \{(\mbx_i, y_i)\}_{i=1}^n \) is drawn from an unknown joint distribution \( \mathcal{P} \) over \( \mathcal{X} \times \mathcal{Y} \), with each \( \mbx_i \in \mathbb{R}^m \) and \( \mathbf{y}_i \in \{0,1\}^C \) being the one-hot encoding of label \( y_i \). We consider a probabilistic classifier \( f \colon \mathcal{X} \rightarrow \Delta^C \), where \( \Delta^C \) is the \( (C-1) \)-dimensional probability simplex. Let \( \mathbf{\hat{p}}_i = f(\mbx_i)\) denote the predicted class probabilities for \( \mbx_i \) and let $\gamma$ be the bandwidth parameter used to compute the kernel estimates $\hat{\theta}(\mathbf y \mid \mbx)$ on a disjoint set of instances.

Let a deterministic binning function
\(
\beta:\Delta^C \rightarrow \{1,\ldots,m_B\}
\)
partition the probability simplex \(\Delta^C\) into 
\( m_B \) multidimensional (\cref{appendix:general_assumptions}) disjoint bins \( \{B_b\}_{b=1}^{m_B} \). For each bin \( B_b \), define the index set of points that fall into it as \( I_b = \{ i : \mathbf{\hat{p}}_i \in B_b \} \). Finally, let \( |B_b|\) denote the bin cardinality and their cumulative sums $S_{b} = \sum_{i \le b} |B_i|$. The $ECCE$ is:

\[
 ECCE = \sum_{c=1}^C \sum_{b=1}^{m_B}\pi_c\frac{S_{b}}{n}\Bigg|\sum_{i=1}^b\frac{|B_{i}|}{S_{b}}\frac1{|B_{i}|}\sum_{j \in \mathcal{I}_i} \Bigg(\mathbf{1}\{y_j=c\}-f_c(\mbx_j)\Bigg)\Bigg| \notag
\] 
And, with at least probability $1-\delta \in [0,1]$, a bound of similar form of the one of Theorem~\ref{thm:generic_bound} applies:
 \[
 \Pr\Bigg(ECCE \le \sum_{c=1}^C\sum_{b=1}^{m_B} \pi_c\cdot w_b \Bigg(\varepsilon +  \sqrt{ \frac{ \log(\frac{2Cm_B}{\delta}) }{ 2|\Psi(b,c)| } }\Bigg)\Bigg) \geq 1-\delta
\]
\begin{proof} We rewrite class-wise $ECCE$ with the use of per-instance kernel estimates \(\hat{\theta}_c(\mathbf{y}_i\mid \mbx_i)\):
\begin{align*}
& \sum_{c=1}^C \sum_{b=1}^{m_B}\pi_c\frac{S_b}{n}\Bigg|\sum_{i=1}^b\frac{|B_i|}{S_{b}}\frac1{|B_i|}\sum_{j \in \mathcal{I}_i} \Bigg(\mathbf{1}\{y_j=c\}-f_c(\mbx_j)\Bigg)\Bigg| \\ = & \sum_{c=1}^C \sum_{b=1}^{m_B}\pi_c\frac{S_b}{n}\Bigg|\sum_{i=1}^b\frac{|B_i|}{S_{b}}\frac1{|B_i|}\sum_{j \in \mathcal{I}_i} \Bigg(\mathbf{1}\{y_j=c\} - \hat{\theta}_c(\mathbf y_j \mid \mbx_j) + \hat{\theta}_c(\mathbf y_j \mid \mbx_j) - f_c(\mbx_j)\Bigg)\Bigg| \\ 
&\le \sum_{c=1}^C\sum_{b=1}^{m_B} \pi_c\frac{S_b}{n}\Bigg[\underbrace{\Bigg|\sum_{i=1}^b\frac{|B_i|}{S_{b}}\frac1{|B_i|}\sum_{j \in \mathcal{I}_i} \Bigg(\mathbf{1}\{y_j=c\} - \hat{\theta}_c(\mathbf y_j \mid \mbx_j) \Bigg) \Bigg|}_{\text{empirical fluctuation}} + \underbrace{\Bigg| \sum_{i=1}^b\frac{|B_i|}{S_{b}}\frac1{|B_i|}\sum_{j \in \mathcal{I}_i} \Bigg( \hat{\theta}_c(\mathbf y_j \mid \mbx_j) - f_c(\mbx_j)\Bigg)\Bigg|}_{\text{miscalibration}}\Bigg].
\notag
\end{align*}

Recall that by the local calibration assumption, for every instance \(i\) we have
\(\| f(\mbx_i) - \hat{\theta}(\mathbf{y}_i\mid \mbx_i)\|_1 \le \varepsilon\). Consequently each coordinate satisfies
\(| f_c(\mbx_i)-\hat{\theta}_c(\mathbf{y}_i\mid \mbx_i)| \le \varepsilon\),  for the \textbf{miscalibration} component:
\[
\Bigg| \sum_{i=1}^b\frac{|B_i|}{S_{b}}\frac1{|B_i|}\sum_{j \in \mathcal{I}_i} \Bigg( \hat{\theta}_c(\mathbf y_j \mid \mbx_j) - f_c(\mbx_j)\Bigg)\Bigg| \le \varepsilon. \notag
\]

In the following we apply Hoeffding inequality to bound the \textbf{Empirical fluctuation} term, we clarify the underlying assumptions our bound. More precisely, to apply Hoeffding, we need independent, bounded, zero-mean summands. 

Under the \textbf{conditioning assumptions} (Appendix~\ref{appendix:conditioning_ass}), the only source of randomness in the term
\( \frac{1}{|B_i|}\sum_{j\in I_i}\big(\mathbf{1}\{y_j=c\}-\hat{\theta}_c(\mathbf{y}_j\mid \mbx_j)\big) \notag
\)
is the label variables $\{y_j\}_{j \in I_i}$. For each $j$, the summand satisfies:  

\begin{enumerate}
    \item \textbf{Zero mean:}
    \[
    \mathbb{E}\!\left[\mathbf{1}\{y_j=c\} - \hat{\theta}_c(\mathbf{y}_j\mid \mbx_j) \;\middle|\; \mbx_j\right] 
    = P(y_j=c \mid \mbx_j) - \hat{\theta}_c(\mathbf{y}_j\mid \mbx_j) \approx 0, \notag
    \]
    and exactly zero if the estimator is consistent in mean absolute error (MAE) (refer to Appendix~\ref{appendix:kernel_assump} for all the details).  
    \item \textbf{Independence:} the pairs $(\mbx_j,y_j)$ are i.i.d., and conditioning on the $\mbx_j$ leaves the labels $\{y_j\}$ independent.  
    \item \textbf{Boundedness:} both $\mathbf{1}\{y_j=c\}$ and $\hat{\theta}_c(\mathbf{y}_j\mid \mbx_j)$ lie in $[0,1]$, hence their difference lies in $[-1,1]$.  
\end{enumerate}

Therefore, the summands are independent, bounded in $[-1,1]$, and zero-mean. Hoeffding’s inequality applies to their average, yielding the desired concentration bound. Choosing \(\tau_b=\sqrt{\tfrac{\log(\frac{2Cm_B}{\delta})}{2 |B_i|}}\) and applying the union bound over the \(m_B\cdot C\) bin–class pairs yields: with probability at least \(1-\delta\),
\[
\forall b,c:\qquad \frac{1}{|B_i|}\sum_{j\in I_i}\big(\mathbf{1}\{y_j=c\}-\hat{\theta}_c(\mathbf{y}_j\mid \mbx_j)\big)
\le \sqrt{\frac{\log(\frac{Cm_B}{\delta})}{2 |B_i|}}.
\notag
\]
Then with high probability:
\begin{align*}
&\sum_{c=1}^C \sum_{b=1}^{m_B}\pi_c\frac{S_b}{n}\Bigg[\underbrace{\Bigg|\sum_{i=1}^b\frac{|B_i|}{S_{b}}\frac1{|B_i|}\sum_{j \in \mathcal{I}_i} \Bigg(\mathbf{1}\{y_j=c\} - \hat{\theta}_c(\mathbf y_j \mid \mbx_j) \Bigg) \Bigg|}_{\text{empirical fluctuation}} + \underbrace{\Bigg| \sum_{i=1}^b\frac{|B_i|}{S_{b}}\frac1{|B_i|}\sum_{j \in \mathcal{I}_i} \Bigg( \hat{\theta}_c(\mathbf y_j \mid \mbx_j) - f_c(\mbx_j)\Bigg)\Bigg|}_{\text{miscalibration}}\Bigg] \\
&\le \sum_{c=1}^C \sum_{b=1}^{m_B}\pi_c\frac{S_b}{n}\Bigg[\sum_{i=1}^b\frac{|B_i|}{S_{b}}\sqrt{\frac{\log(\frac{Cm_B}{\delta})}{2 |B_i|}} + \sum_{i=1}^b\frac{|B_i|}{S_{b}}\varepsilon\Bigg]
\end{align*}
Since $\sum_{i=1}^b\nicefrac{|B_i|}{S_b} = 1$ , the bound then simplifies as follows: 
\[
\Pr\Bigg( ECCE \le  \sum_{c=1}^C \sum_{b=1}^{m_B} \pi_c\cdot w_b\Bigg(\varepsilon +\sqrt{\frac{\log(\frac{Cm_B}{\delta})}{2 |\Psi(b,c)|}}\Bigg) \Bigg) \ge 1-\delta \notag
\] 
with \(\Psi(b,c) = B_{i^\star}\) with \( i^\star = \arg \min_{i \le b} |B_i|\) and \(w_b = \nicefrac{S_b}{n}\). 
\paragraph{Class-wise metric.} 
In the context of class-wise metrics that use confidence based binning, the only difference is the introduction of a dependence between bins' cardinalities and the classes: $|B_b| \to |B_{b,c}|$. Then, with probability at least $1-\delta$, the bound applies with the same exact form under the following minor changes: 
\[
\textit{class-wise } ECCE \le 
\sum_{c=1}^C\sum_{b=1}^{m_B} \pi_c\frac{S_{b,c}}{n}  \Bigg(\varepsilon + \sqrt{\frac{\log(\frac{Cm_B}{\delta})}{2 \min_{i}|B_{i,c}|}} \Bigg)
\]
and setting \(\Psi(b,c) = B_{i^\star,c}\) with \( i^\star=\arg\min_{i}|B_{i,c}|\) and \(w_{b,c}=\nicefrac{S_{b,c}}{n}\) we obtain the bound of \cref{thm:generic_bound}.
\end{proof}
\subsection{Local Calibration and Proximity Bias}
\label{sec:loc_and_prox}
\textit{Proximity bias} is a well-documented phenomenon in probabilistic classifiers \citep{DBLP:conf/nips/XiongDKW0XH23}, where models tend to exhibit systematic miscalibration on sparsely represented instances. This behavior is particularly concerning, as it can introduce unintended biases against underrepresented subpopulations. 
Addressing \textit{proximity bias} is therefore critical to ensuring fairness in algorithmic decision-making, especially in high-stakes domains such as law and medicine, where equitable and reliable predictions are essential. 
The most effective way to characterize this phenomenon is by directly comparing the class frequency distributions of two subgroups that share similar model confidence scores but differ in input-space density. 

We leverage this approach to examine how \textit{local calibration} may mitigate \textit{proximity bias}. 
Specifically, we provide a theoretical decomposition of the change in class frequencies when transitioning from high-density to low-density regions and use this framework to derive a probabilistic upper bound on \textit{proximity bias} under the assumption of \textit{local calibration}.
More precisely, the total error can be decomposed into three components: a stochastic fluctuation, a calibration error and a distribution shift term respectively. The latter captures the extent to which the score distributions vary across different regions of the input space—particularly when transitioning from densely to sparsely represented instances.
%

\begin{theorem}[Error Decomposition of Proximity Bias]
\label{thm:prox_err_dec}
Let \( S_1 \) and \( S_2 \) be two proximity-based sub-bins drawn from the same score-based bin, with cardinalities \( |S_1| \) and \( |S_2| \). Define:
\[
\text{freq}(S_s) := \frac{1}{|S_{s}|} \sum_{i \in S_s} \mathbf{y}_i, \quad
\text{conf}(S_s) := \frac{1}{|S_{s}|} \sum_{i \in S_s} f(\mbx_i)\notag
\]
Let $f$ be locally calibrated with error $\varepsilon$ with respect to kernel estimates on an arbitrarily large auxiliary dataset, then conditioning on the evaluation samples $\{\mbx_i\}_{i=1}^n$ and the corresponding consistent kernel estimates, with probability at least \( 1 - \delta \in [0,1] \) the difference in class frequencies between the two sub-bins is bounded as follows:
\[
\Pr\left( \left\| \text{freq}(S_1) - \text{freq}(S_2) \right\|_1 
\leq 2\varepsilon +\sqrt{\frac{2\log(\frac{4C}{\delta})}{\min(|S_1|, |S_2|)}} + \left\| \text{conf}(S_1) - \text{conf}(S_2) \right\|_1 \right) \geq 1 - \delta.
\]
\end{theorem}

\begin{proof}
Suppose the simplex \( \Delta^C \) is partitioned into \( m_B \) score-based disjoint bins \( \{B_b\}_{b=1}^{m_B} \). 
Each score-based bin is further subdivided by grouping points with similar feature-space proximity. For each point \( \mbx_i \), define its proximity score:
\[
\pi_k(\mbx_i) := \frac{1}{k} \sum_{j=1}^{k} \|\phi(\mbx_i) - \phi(\mbx_{(i,j)})\|_2, \notag
\]
where \( \mbx_{(i,1)}, \ldots, \mbx_{(i,k)} \) are the \( k \) nearest neighbors of \( \mbx_i \) in feature space (excluding \( \mbx_i \) itself).
%

%
We begin by applying the triangle inequality:
\begin{align}
\left\| \text{freq}(S_1) - \text{freq}(S_2) \right\|_1 = &  \left\| \Big(\text{freq}(S_1) - \text{conf}(S_1)\Big)
+   \Big(\text{conf}(S_1) - \text{conf}(S_2) \Big) 
+  \Big(\text{conf}(S_2) - \text{freq}(S_2) \Big) \right\|_1 \notag\\
&\leq \left\| \Big(\text{freq}(S_1) - \text{conf}(S_1)\Big) + \Big(\text{conf}(S_1) - \text{conf}(S_2)\Big) \right\|_1 
+ \left\| \text{conf}(S_2) - \text{freq}(S_2) \right\|_1 \notag\\
&\leq \left\| \text{freq}(S_1) - \text{conf}(S_1) \right\|_1 
+ \left\| \text{conf}(S_1) - \text{conf}(S_2) \right\|_1 
+ \left\| \text{conf}(S_2) - \text{freq}(S_2) \right\|_1.
\label{eq:triangle}
\end{align}

From Theorem~\ref{thm:generic_bound}, which applies identically to any bin or subset under the same \textbf{conditioning assumptions} of Appendix~\ref{appendix:conditioning_ass}, we have that for any $\delta'\in[0,1]$ the following probabilistic bound for each sub-bin \( S_s \):  
\[
\Pr\left( \left\| \text{freq}(S_s) - \text{conf}(S_s) \right\|_1 \leq \varepsilon + \sqrt{\frac{\log(\frac{2C}{\delta'})}{2|S_s|}} \right) \geq 1 - \delta'. \notag
\]

Now define the following events:
\begin{align*}
A &:= \left\{ \left\| \text{freq}(S_1) - \text{conf}(S_1) \right\|_1 > \varepsilon + \eta_{\delta_1} \right\}, \quad \text{where } \eta_{\delta_1} := \sqrt{\frac{\log(\frac{2C}{\delta'})}{2|S_1|}}, \\
B &:= \left\{ \left\| \text{freq}(S_2) - \text{conf}(S_2) \right\|_1 > \varepsilon + \eta_{\delta_2} \right\}, \quad \text{where } \eta_{\delta_2} := \sqrt{\frac{\log(\frac{2C}{\delta'})}{2|S_2|}}.
\end{align*}

Applying the union bound:
\[
\Pr(A \cup B) \leq \Pr(A) + \Pr(B) \leq 2\delta'. \notag
\]

Thus, with probability at least \( 1 - 2\delta' \), both events do not occur:
\[
\Pr(\neg A \cap \neg B) \geq 1 - 2\delta'. \notag
\]

Under this event, we can bound \cref{eq:triangle}. More precisely, both the first and third terms is bounded by \( \varepsilon + \sqrt{\frac{\log(\frac{2C}{\delta'})}{2n_{S_s}}} \), and by choosing $\delta'=\frac{\delta}{2}$ we can conclude:
\[
\Pr\left( \left\| \text{freq}(S_1) - \text{freq}(S_2) \right\|_1 
\leq 2\varepsilon +\sqrt{\frac{2\log(\frac{4C}{\delta})}{\min(|S_1|, |S_2|)}} + \left\| \text{conf}(S_1) - \text{conf}(S_2) \right\|_1 \right) \geq 1 - \delta.
\] \end{proof}

Intuitively, under local calibration, predicted scores approximate true class frequencies. Therefore, any shift in the score distribution within a bin implies a corresponding shift in the underlying class frequencies. The error due to this distributional inconsistency can be reduced by refining the density-based bins, but finer binning leads to smaller sample sizes per bin, thereby increasing the stochastic fluctuation error. This trade-off highlights an inherent tension in binning procedures: reducing distribution shift comes at the cost of increased variance. 

Although limited availability of data is problematic in capturing \textit{proximity bias} of a \textit{locally calibrated} model, we can investigate the phenomenon from a theoretical perspective in the limit of infinite data availability. This allows us to schedule the bin width reduction, bounding the admissible score change, while keeping a sufficient bin cardinality to workaround the inherent trade-off between the two. This analysis allows to conclude that, under \textit{local calibration}, the value of \textit{proximity bias}, if it could be computed with access to infinite data, would be tightly bounded by the model’s calibration error, which is explicitly controlled by the \textit{local calibration} property.

\begin{corollary}[Infinite Limit of Proximity bias under Local Calibration]

Let the assumptions of Theorem~\ref{thm:prox_err_dec} hold, and assume moreover that the conditional density \(h(\mbx\mid\mathbf{\hat{p}})\) and marginal density \(q_f(\mathbf{\hat{p}})\) be continuous in a neighborhood of \((\mbx_i, \mathbf{\hat{p}_i})\), with \(h(\mbx_i\mid\mathbf{\hat{p}_i}) > 0\) and \(q_f(\mathbf{\hat{p}_i}) > 0\) for some $i \in S_s$. Then, the bound on the proximity bias asymptotically satisfies:

\[
\forall\; \xi > 0: \lim_{n\to\infty}\Pr\Big(\| \mathrm{freq}(S_1) - \mathrm{freq}(S_2)\|_1 \le 2\varepsilon + \xi \Big) = 1
\]

\end{corollary}
Thus, the empirical difference in class frequencies between proximity sub-bins is controlled by \( \varepsilon \) up to vanishing stochastic fluctuations. This theoretical result allows to infer that \textit{proximity bias} is directly controlled by the \textit{local calibration} property of a model.
\begin{proof}
Let the setting and notation be as in Theorem~\ref{thm:prox_err_dec}. Additionally, fix a target confidence vector \(\mathbf{\hat{p}}_0 \in (0,1)^C\) and let the confidence (scores) bin centered at \(\mathbf{\hat{p}}_0\) be:
\[
B_n(\mathbf{\hat{p}}_0) = \{\,\mathbf{\hat{p}} : |\mathbf{\hat{p}} - \mathbf{\hat{p}}_0| \le w_n\,\}, \notag
\]
with radius \(w_n \to 0\).
Define the set of indices of samples in this bin as:
\[
\mathcal{I}_n(\mathbf{\hat{p}}_0) = \{\,i : \mathbf{\hat{p}}_i \in B_n(\mathbf{\hat{p}}_0)\,\}. \notag
\]
Within this set, consider two disjoint density-based sub-bins 
\(S_1, S_2 \subseteq \mathcal{I}_n(\mathbf{\hat{p}}_0)\) corresponding to local neighborhoods in feature space. 
Additionally, let the conditional density \(h(\mbx\mid\mathbf{\hat{p}})\) and marginal density \(q_f(\mathbf{\hat{p}})\) be continuous in a neighborhood of \((\mbx_i, \mathbf{\hat{p}_i})\), with \(h(\mbx_i\mid\mathbf{\hat{p}_i}) > 0\) and \(q_f(\mathbf{\hat{p}_i}) > 0\) for some $i \in S_s$.
  
 By the triangle inequality (as in \eqref{eq:triangle}),
\[
\|\mathrm{freq}(S_1)-\mathrm{freq}(S_2)\|_1
\le \|\mathrm{freq}(S_1)-\mathrm{conf}(S_1)\|_1
+ \|\mathrm{conf}(S_1)-\mathrm{conf}(S_2)\|_1
+ \|\mathrm{conf}(S_2)-\mathrm{freq}(S_2)\|_1.
\notag\]
From Theorem~\ref{thm:prox_err_dec} (coordinate-wise Hoeffding + union bound) we have, for any fixed \(\delta=2\cdot\delta'\),
\[
\Pr\!\Bigg(\|\mathrm{freq}(S_s)-\mathrm{conf}(S_s)\|_1 \le \varepsilon + \sqrt{\frac{\log(\frac{2C}{\delta'})}{2\min(|S_1|, |S_2|)}}\Bigg) \;\ge\; 1-\delta' \notag
\]
Applying the union bound gives that with probability at least \(1-\delta\) both deviations are bounded simultaneously:
\[
\left\| \text{freq}(S_1) - \text{freq}(S_2) \right\|_1 
\leq 2\varepsilon +\sqrt{\frac{2\log(\frac{4C}{\delta})}{\min(|S_1|, |S_2|)}} + \left\| \text{conf}(S_1) - \text{conf}(S_2) \right\|_1 
\notag\]
Then, under the local calibration assumption
\(\|f(\mbx_j)-\hat{\theta}(\mathbf y_j\mid \mbx_j)\|_1\le\varepsilon\) for all \(j \in D\), we will prove that:
\[
\forall\; \xi > 0: \lim_{n\to \infty}\Pr\Big(\| \mathrm{freq}(S_1) - \mathrm{freq}(S_2)\|_1 \le 2\varepsilon + \xi \Big) = 1
\]

\vspace{0.2cm}
\noindent\textbf{Step 1. Shrinking confidence bins controls differences in predicted scores.}  
By construction, for every sample \(j \in \mathcal{I}_n(\mathbf{\hat{p}}_0)\),
\[
\mathbf{\hat{p}}_j \in [\mathbf{\hat{p}}_0 - w_n,\, \mathbf{\hat{p}}_0 + w_n]. \notag
\]

and thus, for any two sub-bins \(S_1, S_2 \subseteq \mathcal{I}_n(\mathbf{\hat{p}}_0)\),
\[
\left\|\mathrm{conf}(S_1) - \mathrm{conf}(S_2)\right\|_1 \le 2Cw_n. \notag
\]
Therefore, as \(w_n \to 0\), the difference in average predicted confidences between any two density-based sub-bins within the same score bin also vanishes:
\[
\left\|\mathrm{conf}(S_1) - \mathrm{conf}(S_2)\right\|_1 \xrightarrow{n\to\infty} 0.
\]

\vspace{0.2cm}
\noindent\textbf{Step 2. Maintaining infinite data within shrinking bins.}  
We now show that it is possible to shrink both the confidence-bin width \(2w_n\) and the density sub-bin radius \(r_n\) simultaneously, while guaranteeing that each sub-bin still contains infinitely many samples with high probability. As a consequence the stochastic square root term asymptotically vanishes in probability.

Fix a ball \(B(\mathbf{x}_i, r_n) \subseteq \mathcal{R}\) centered in $\mbx_i$ and with volume \(\mathrm{vol}(B(0,r_n))\) where \(\mathcal{R}\) is the region of space associated to a sub-bin. Likewise, fix a ball $B_n(\mathbf{\hat{p}}_i, w_i) \subseteq B_n(\mathbf{\hat{p}}_0)$. 
The joint probability that a sample lies in both balls is:
\[
\Pr\!\big(\mathbf{X} \in B(\mathbf{x}_i, r_n),\, f(\mathbf{x}) \in B_n(\mathbf{\hat{p}}_i, w_i)\big)
= \int_{B_n(\mathbf{\hat{p}}_i, w_i)} \!\!\int_{B(\mathbf{x}_i, r_n)} h(\mathbf{x}\mid \mathbf{\hat{p}})\, q_f(\mathbf{\hat{p}})\, d\mathbf{x}\, d\mathbf{\hat{p}}. \notag
\]
Then, by the Lebesgue differentiation theorem 
(see e.g.\ Theorem~1.3 in \citep{stein2009real}),
for sufficiently small \(r_n\) and \(w_i\),
\[
\Pr\!\big(\mathbf{X} \in B(\mathbf{x}_i, r_n),\, f(\mathbf{X}) \in B_n(\mathbf{\hat{p}}_i, w_i)\big)
\;=\;
h(\mathbf{x}_i\mid \mathbf{\hat{p}}_i)\,q_f(\mathbf{\hat{p}}_i)\,
\mathrm{vol}(B(0,r_n))\,\mathrm{vol}(B(0,w_i)) + o(r_n^d w_i^{C-1}), \notag
\]
where \(o(r_n^d w_i^{C-1})\) denotes a term that becomes negligible compared to the product of the volumes \(r_n^d w_i^{C-1}\). 
Note that this result leverages the norm equivalence for finite-dimensional spaces like $\Delta^{C} \subset \mathbb{R}^{C-1}$. As a consequence, using $L_1$ or $L_2$ balls only changes bounds by constant factors which do not affect asymptotic rates. 
Moreover, by the \((\varepsilon,\delta)\)-definition of continuity (Weiserstrass-Jordan), there exist finite positive constants $c_h,C_h,q_{\min},q_{\max}$ and a neighborhood $U$ of $(\mathbf{x}_i,\hat{\mathbf p}_i)$ such that:
\[
0 < c_h \le h(\mathbf{x}\mid\hat{\mathbf p}) \le C_h < \infty,\qquad
0 < q_{\min} \le q_f(\hat{\mathbf p}) \le q_{\max} < \infty,
\qquad \forall (\mathbf{x},\hat{\mathbf p})\in U. \notag
\]
For all sufficiently small $r_n,w_i$, so that $B(\mathbf{x}_i,r_n)\times B_n(\mathbf{\hat{p}}_i, w_i)\subset U$, the joint probability admits a two-sided bound:
\[
c_h q_{\min}\,\mathrm{vol}(B(0,r_n))\,\mathrm{vol}(B(0,w_i))
\;\le\;
\Pr\!\big(\mathbf{X}\!\in\! B(\mathbf{x}_i,r_n),\, f(\mathbf{X})\!\in\! B_n(\hat{\mathbf p}_0)\big)
\;\le\;
C_h q_{\max}\,\mathrm{vol}(B(0,r_n))\,\mathrm{vol}(B(0,w_i)) . \notag
\]
Hence the expected number of points in a sub-bin satisfies:
\[
\mathbb{E}[|S_s|] \asymp n\,r_n^{d}\,w_i^{C-1}, \notag
\]
where \(\asymp\) indicates asymptotic proportionality, meaning that \(E[|S_s|]\) grows at the same rate as \(nr_n^dw_i^{C-1}\) up to constant factors. 
Choosing sequences \(w_i = n^{-\alpha}\) and \(r_n = n^{-\beta}\) with: 
\[
0 < \alpha < \frac1{C-1}, \qquad 0 < \beta < \frac{1-\alpha(C-1)}{d}, \notag
\]
we obtain: 
\[
n\,r_n^{d}\,w_i^{C-1} = n^{\,1 - \alpha(C-1) - \beta d} \to \infty, \notag
\]
while both \(w_i, r_n\to 0\).
Since each sample $\{(\mathbf{x}_j, f(\mathbf{x}_j))\}_{j=1}^n$ is drawn i.i.d.,
the number of samples in a local sub-bin $S_s$ follows a binomial distribution 
$|S_s| \sim \mathrm{Binomial}(n, p_n)$ with success probability 
$p_n = \Pr\!\big((\mathbf{x}, f(\mathbf{x})) \in B(\mathbf{x}_i, r_n) \times B_n(\mathbf{\hat{p}}_i, w_i)\big)$. As a consequence, for any $\eta \in (0,1)$, the Chernoff bound gives:
\[
\Pr\!\big(|S_s| \geq (1-\eta)\,\mathbb{E}[|S_s|]\big)
\geq 1 -\exp\!\Big(-\tfrac{\eta^2}{2}\,\mathbb{E}[|S_s|]\Big). \notag
\]
Since $\mathbb{E}[|S_s|]\to\infty$ as $n\to\infty$, with probability tending to one:
\[
|S_s| \ge (1-\eta)\,\mathbb{E}[|S_s|] \to \infty
\]
Which allows us to conclude:
\[
\forall\; \xi > 0: \lim_{n\to \infty}\Pr\Big(\| \mathrm{freq}(S_1) - \mathrm{freq}(S_2)\|_1 \le 2\varepsilon + \xi \Big)= 1
\]
\end{proof}
\paragraph{Remark.}
The argument readily extends to the case where each proximity-based sub-bin \(S_s\) 
is a finite union of disjoint regions \(\{R_{s,k}\}_{k=1}^{K_s}\).
Under the same regularity and consistency assumptions applied component-wise 
(shrinking diameters and diverging per-component sample sizes), 
the concentration and continuity arguments hold uniformly over components, 
and the aggregate deviation remains bounded by \(2\varepsilon\) up to vanishing terms with maximum probability.
We restrict the proof to a single region per sub-bin for notational simplicity.
\subsection{Illustrative Example}

In this section, we present a toy example to highlight potential pitfalls of density-based calibration. Specifically, we show that the choice of bin width plays a critical role: overly wide bins may lead to ineffective recalibration, while overly fine bins require large sample sizes and can become computationally prohibitive. Our goal here is to raise awareness on risks that can arise in practice. 

\begin{figure}[ht]
    \centering
    \includegraphics[width=0.8\linewidth]{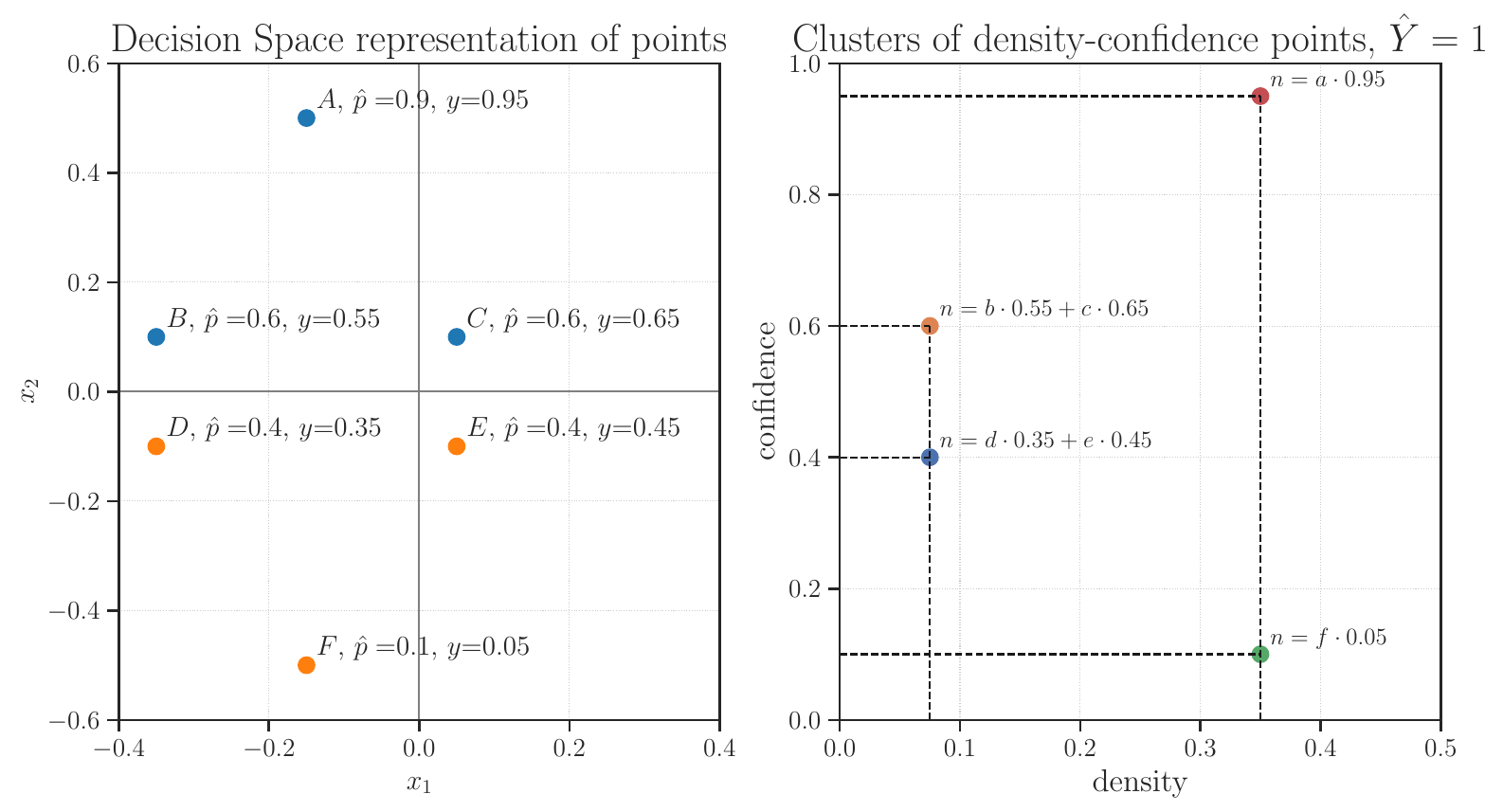}
    \caption{Points in the decision space (right) and their mapping to density-confidence space for calibration (left).}
    \label{fig:your-label}
\end{figure}

Consider a binary probabilistic classifier \( f(\cdot) \) and a dataset \( D = \{(\mbx_i, y_i)\}_{i=1}^n \), where each input \( \mbx_i \in \mathbb{R}^m\) takes one of six distinct sets of values (here $m=2$ for visualization purposes). Figure~\ref{fig:your-label} provides a visual representation of these points in the decision space (left), as well as their corresponding locations in the density-confidence space used for calibration (right).

Grouping points by their coordinates yields six disjoint regions. Within each region, the classifier assigns a constant predicted probability (all inputs have same values within region). Table~\ref{tab:first_table} reports the size, density, predicted probability, and empirical label frequency for each region.

\begin{table}[H]
    \centering
    \caption{The six disjoint regions characterized by density, predicted probability, and empirical frequency.}
    \label{tab:first_table}
    \begin{tabular}{|c|c|c|c|c|}
\hline
\textbf{Set} & \textbf{Size} & \textbf{Density} & \textbf{p} & \textbf{Y} \\
\hline
A & $a$ & $0.35$ & $0.9$ & $0.95$ \\
\hline
B & $b$ & $0.075$ & $0.6$ & $0.55$  \\
\hline
C & $c$ & $0.075$ & $0.6$ & $0.65$  \\
\hline
D & $d$ & $0.075$ & $0.4$ & $0.35$  \\
\hline
E & $e$ & $0.075$ & $0.4$ & $0.45$  \\
\hline
F & $f$ & $0.35$ & $0.1$ & $0.05$  \\
\hline
\end{tabular}
\end{table}

In Figure~\ref{fig:your-label} (right), each point is plotted according to its predicted probability and estimated local density. A most fine-grained approach in calibration is to aggregate predictions based on proximity in this joint space. That is, the calibrated probability for a point is conditioned on both its score \( \hat{P} \) and density estimate \( \hat{D} \), and is computed as:

\[
p_{\text{cal}} = \Pr(\hat{Y} = Y \mid \hat{P}, \hat{D}) = \frac{\Pr(\hat{P}, \hat{D} \mid \hat{Y} = Y) \cdot \Pr(\hat{Y} = Y)}{\Pr(\hat{P}, \hat{D} \mid \hat{Y} = Y) \cdot \Pr(\hat{Y} = Y) + \Pr(\hat{P}, \hat{D} \mid \hat{Y} \neq Y) \cdot \Pr(\hat{Y} \neq Y)} \notag
\]

We focus on calibrating predictions for points with \( \hat{P} = 0.6 \) and \( \hat{D} = 0.075 \). Among the six regions,  regions \( b \) and \( c \) match this pair of values. Then:

\begin{align*}
\Pr(\hat{P}=0.6, \hat{D}=0.075 \mid \hat{Y}=1) &= \frac{b \cdot 0.55 + c \cdot 0.65}{a \cdot 0.95 + f \cdot 0.05 + d \cdot 0.35 + e \cdot 0.45 + b \cdot 0.55 + c \cdot 0.65} = \frac{\text{NUM}_1}{\text{DEN}_1} \notag \\
\Pr(\hat{P}=0.6, \hat{D}=0.075 \mid \hat{Y}=0) &= \frac{b \cdot 0.45 + c \cdot 0.35}{a \cdot 0.05 + f \cdot 0.95 + d \cdot 0.65 + e \cdot 0.55 + b \cdot 0.45 + c \cdot 0.35} = \frac{\text{NUM}_2}{\text{DEN}_2} \notag
\end{align*}

Since
\[
\frac{\Pr(\hat{Y} \neq Y)}{\Pr(\hat{Y} = Y)} = \frac{\text{DEN}_2}{\text{DEN}_1}, \notag
\]
we simplify the calibrated probability as:

\[
p_{\text{cal}} = \frac{\text{NUM}_1}{\text{NUM}_1 + \text{NUM}_2} = \frac{b \cdot 0.55 + c \cdot 0.65}{b + c} \notag
\]

Since regions \( b \) and \( c \) have similar sizes (i.e., \( b \approx c \)), then:

\[
p_{\text{cal}} \approx \frac{0.55 + 0.65}{2} = 0.6 \neq [0.55, 0.65] \notag
\]

This recalibrated probability equals a weighted average of the empirical frequencies of regions $b$ and $c$. As such, it cannot simultaneously correct both, and whichever side is under/overconfident remains miscalibrated after recalibration. The magnitude of miscalibration aggravates when one region is underconfident and the other is overconfident. Such heterogeneous calibration errors do not exclusively occur when regions with similar predicted confidences differ in density but can also arise in presence of differences in class overlap or representation smoothness, as commonly observed in deep neural networks \citep{DBLP:conf/icml/GuoPSW17,DBLP:conf/nips/KullPKFSF19}. Consequently, grouping them within the same density–confidence bin averages incompatible local behaviors, masking or amplifying miscalibration. While narrower bins could mitigate this effect, they quickly become sample-inefficient and computationally demanding.
\section{Experimental Details}
\label{app:experimental_details}

In this section, we provide all the details that should allow exact reproducibility of our results.
\subsection{Training of Classifiers}
We report here all training configurations for the classifiers used in the calibration experiments. All classifiers used categorical cross-entropy, and no batch normalization layers or weight decay were applied during fine-tuning.

\paragraph{\texttt{CIFAR-10.}} 
We use a \texttt{ResNet-50} architecture initialized with \texttt{IMAGENET1K\_V2} pre-trained weights. 
A dropout layer with rate $0.2$ is appended to the final backbone layer, followed by a linear classification head. 
During fine-tuning, all layers are frozen except for the last backbone block and the classification head. 
Optimization is performed for $9$ epochs using the \texttt{Adam} optimizer~\citep{DBLP:journals/corr/KingmaB14} with a learning rate of $3\!\times\!10^{-4}$.

\paragraph{\texttt{CIFAR-100.}} 
We adopt a \texttt{ResNet-152} model pre-trained on \texttt{IMAGENET1K\_V2}. 
A dropout layer with a rate $0.5$ is inserted before the classification layer. 
All layers except the last backbone block and the classifier are frozen during training. 
We optimize for $9$ epochs using \texttt{Adam} with a learning rate of $3\!\times\!10^{-4}$.

\paragraph{\texttt{TissueMNIST.}} 
We employ a \texttt{ResNet-50} backbone initialized with \texttt{IMAGENET1K\_V2} weights. 
A dropout layer with a rate $0.2$ is applied before the linear classification layer. 
As in the previous setups, all layers except the last backbone block and the classification layer are trainable. 
We train for $10$ epochs using the \texttt{Adam} optimizer with a learning rate of $3\!\times\!10^{-4}$.
\subsection{Training of LoCal Nets}
We report here all training configurations for \LCN{} in the calibration experiments.

\paragraph{Residual Modelling.}
The \LCN{} operates in a residual fashion. 
Given the intermediate representations $\phi(\mathbf{x})$ extracted from a pre-trained backbone, let $\phi_{\mathrm{PCA}}(\mathbf{x})$ denote the reduced feature representation obtained via Principal Component Analysis (PCA). 
The \LCN{} processes $\phi(\mathbf{x})$ through its hidden layer to produce refined features $\tilde{\phi}_{\mathrm{PCA}}(\mathbf{x})$ and logits $\tilde{g}(\mathbf{x})$. 
The final representations $\phi_{\mathrm{PCA}}'(\mathbf{x})$ and $g'(\mathbf{x})$ are obtained through a weighted residual combination:
\[
\phi_{\mathrm{PCA}}'(\mathbf{x}) = \tilde{\phi}_{\mathrm{PCA}}(\mathbf{x}) + w_{\phi} \cdot \phi_{\mathrm{PCA}}(\mathbf{x}) + b_\phi, 
 \quad  
 g'(\mathbf{x}) = \tilde{g}(\mathbf{x}) + w_{g} \cdot g(\mathbf{x}) + b_g,
\]
where $w_{\phi}, b_\phi, w_g$ and $b_g$ are learnable scalar weights and biases that adaptively control the contribution of the original features and logits, respectively. Weights are initialized as $1$ and biases are randomly sampled from normal distributions with $0.$ location and $0.01$ scale parameters.
The residual formulation provides strong initialization for \LCN{} outputs, enables preserving the semantic content of the backbone features while introducing locally calibrated corrections, and improves both stability and convergence to meaningful solutions.

\paragraph{\texttt{CIFAR-10.}} 
The \LCN{} is implemented as a fully connected network with a single hidden layer of size $64$ and dropout rate $0.3$. 
It has two output heads: one of dimension $10$ (corresponding to the number of classes) and one of dimension $50$, used for the PCA-reduced feature representations. 
The loss weighting hyperparameter $\lambda$ is set to $1$, ensuring equal contribution of both components of the objective. 
The fixed kernel bandwidth $\gamma$ is set to $10$, chosen to be as small as possible to preserve locality while maintaining stable convergence of the cross-entropy component of the loss, as excessively small values lead to training collapse. 
This choice is validated empirically using a held-out validation set. 
Optimization is performed using the \texttt{Adam} optimizer with learning rate $1\!\times\!10^{-3}$, for $22$ epochs and a batch size of $1024$.

\paragraph{\texttt{CIFAR-100.}} 
The \LCN{} uses a fully connected architecture with one hidden layer of size $128$ and dropout rate $0.5$. 
As before, it has two output heads: one of dimension $100$ (matching the number of classes) and one of dimension $50$ for the PCA-reduced features. 
We set $\lambda = 1$ for equal loss weighting and fix $\gamma = 10$ for consistency with the other datasets. 
In this case, slightly smaller bandwidth values were found feasible, but $\gamma=10$ was retained for coherence across experiments. 
Optimization uses \texttt{Adam} with learning rate $1\!\times\!10^{-3}$, trained for $30$ epochs with a batch size of $1024$.

\paragraph{\texttt{TissueMNIST.}} 
The \LCN{} is a single-hidden-layer fully connected network with hidden dimension $256$ and dropout rate $0.3$. 
It includes two output heads: one of dimension $8$ (the number of classes) and one of dimension $50$ for the PCA-reduced representations. 
The hyperparameter $\lambda$ is set to $1$, and the bandwidth $\gamma$ is fixed at $10$, following the same locality–stability trade-off principle described above, validated via a held-out set. 
Optimization uses \texttt{Adam} with a learning rate of $1\!\times\!10^{-3}$ for $60$ epochs with a batch size of $1024$. 
\subsection{Metrics}
\label{sec:metrics_impl}
In the following, we provide implementation details for the calibration metrics and the associated hyperparameter configurations used in our experiments to allow full reproducibility of our results.

\paragraph{Class-wise Binning Metrics.} 
For both the \textit{Expected Calibration Error} (ECE) and the \textit{Expected Cumulative Calibration Error} (ECCE), we partitioned $f_c(\mbx)$ into $15$ bins based on predicted confidence scores. 
Empty bins, when present, were excluded from the computation. 
Class-wise calibration errors were first computed independently for each class and subsequently aggregated using class-frequency weights estimated from the training data. 
While \texttt{CIFAR-10} and \texttt{CIFAR-100} are both balanced datasets, \texttt{TissueMNIST} exhibits class imbalance, with priors ranging approximately from $0.32$ to $0.04$.

\paragraph{Class-wise Kernel Metrics.} 
\label{sec:class_wise_lce}
We employ two kernel-based calibration metrics: the \textit{multiclass Local Calibration Error} (LCE) and its maximum variant (\textit{MLCE}). 
To extend $LCE$ to the multiclass setting, we adopt a class-wise formulation analogous to that used for binning-based metrics. 
Specifically, $f_c(\mbx)$ is partitioned into $15$ confidence-based bins, and for each fixed class, we use the corresponding bins to identify the neighborhood of each anchor point for kernel estimation. Bins with fewer than 20 elements were discarded to prevent using unstable kernel estimates. 
For each $i \in  D$, the $LCE$ is computed as the absolute difference between kernel-weighted estimates of predicted confidences and empirical labels of instances in the same confidence bin. Per-sample deviations are then averaged, and the values for each class are combined using priors to obtain the final $LCE$.
The kernel bandwidth parameter was set to $\gamma = 10$, consistent with the bandwidth used during the training of the \LCN{}.

\subsection{Hardware and Training Time}
For our experiments, we use a 16-core machine with an AMD Ryzen 9 7950X CPU and 2 NVIDIA GeForce RTX 4090 GDDR6X with 24GB of memory, OS Ubuntu 22.04.4 LTS.

\section{Detailed Results}

\begin{table}[h]
    \centering
    \scriptsize
    \begin{tabular}{c c  c  c  c  c  c  c }
\toprule
                                              Dataset &      Method &                        $LCE\,\downarrow$ &                       $MLCE\,\downarrow$ &                       $ECCE\,\downarrow$ &                        $ECE\,\downarrow$ &                      $ACC\,\uparrow$ &                       $NLL\,\downarrow$ \\
                                                   \\
\midrule
 
\midrule
\multirow{8}{*}{\rotatebox{90}{\texttt{cifar10}}} & \cellcolor{blue!10}$\textsc{LN}$ & \cellcolor{blue!10}   $\mathbf{.0079\pm .0003}$ & \cellcolor{blue!10}   $\mathbf{.6279\pm .0157}$ & \cellcolor{blue!10}            $.0017\pm .0001$ & \cellcolor{blue!10}            $.0048\pm .0003$ & \cellcolor{blue!10}            $.888\pm .002$ & \cellcolor{blue!10}             $.347\pm .002$ \\
                                                   & $\textsc{DC}$ &             $.0099\pm .0003$ &             $.8133\pm .0195$ &    $\mathbf{.0011\pm .0001}$ &    $\mathbf{.0037\pm .0001}$ &             $.890\pm .002$ &     $\mathbf{.332\pm .007}$ \\
                                                   & $\textsc{KO}$ & $\underline{.0086\pm .0003}$ &             $.8190\pm .0183$ & $\underline{.0012\pm .0001}$ & $\underline{.0045\pm .0002}$ & $\underline{.893\pm .001}$ &              $.340\pm .002$ \\
                                                   & $\textsc{KC}$ &             $.0095\pm .0005$ & $\underline{.7923\pm .0269}$ &             $.0015\pm .0003$ &             $.0054\pm .0004$ &    $\mathbf{.893\pm .001}$ &  $\underline{.340\pm .003}$ \\
                                                   & $\textsc{PS}$ &             $.0180\pm .0005$ &             $.8542\pm .0224$ &             $.0014\pm .0001$ &             $.0085\pm .0003$ &             $.884\pm .001$ &              $.466\pm .004$ \\
                                                   & $\textsc{IR}$ &             $.0115\pm .0003$ &             $.7978\pm .0162$ &             $.0014\pm .0002$ &             $.0047\pm .0001$ &             $.884\pm .001$ &              $.364\pm .008$ \\
                                                   & $\textsc{TS}$ &             $.0127\pm .0006$ &             $.8200\pm .0191$ &             $.0036\pm .0008$ &             $.0072\pm .0010$ &             $.884\pm .001$ &              $.362\pm .008$ \\
                                                   & $\textsc{NC}$ &             $.0154\pm .0005$ &             $.8738\pm .0238$ &             $.0065\pm .0004$ &             $.0153\pm .0006$ &             $.884\pm .001$ &              $.494\pm .018$ \\

\midrule
\multirow{8}{*}{\rotatebox{90}{\texttt{cifar100}}} & \cellcolor{blue!10}$\textsc{LN}$ & \cellcolor{blue!10}   $\mathbf{.0024\pm .0001}$ & \cellcolor{blue!10}   $\mathbf{.7022\pm .0070}$ & \cellcolor{blue!10}            $.0007\pm .0001$ & \cellcolor{blue!10}   $\mathbf{.0018\pm .0001}$ & \cellcolor{blue!10}$\underline{.688\pm .001}$ & \cellcolor{blue!10}   $\mathbf{1.125\pm .002}$ \\
                                                   & $\textsc{DC}$ & $\underline{.0027\pm .0001}$ &             $.8117\pm .0121$ &    $\mathbf{.0007\pm .0001}$ & $\underline{.0019\pm .0001}$ &    $\mathbf{.690\pm .002}$ & $\underline{1.154\pm .009}$ \\
                                                   & $\textsc{KO}$ &             $.0043\pm .0001$ &             $.7575\pm .0101$ &             $.0013\pm .0001$ &             $.0044\pm .0001$ &             $.679\pm .003$ &             $1.352\pm .009$ \\
                                                   & $\textsc{KC}$ &             $.0043\pm .0001$ &             $.7594\pm .0055$ &             $.0013\pm .0001$ &             $.0044\pm .0002$ &             $.679\pm .004$ &             $1.351\pm .007$ \\
                                                   & $\textsc{PS}$ &             $.0053\pm .0001$ & $\underline{.7031\pm .0409}$ &             $.0011\pm .0001$ &             $.0033\pm .0001$ &             $.670\pm .002$ &             $1.618\pm .007$ \\
                                                   & $\textsc{IR}$ &             $.0031\pm .0001$ &             $.8049\pm .0116$ & $\underline{.0007\pm .0001}$ &             $.0020\pm .0001$ &             $.670\pm .002$ &             $1.437\pm .029$ \\
                                                   & $\textsc{TS}$ &             $.0034\pm .0001$ &             $.8239\pm .0173$ &             $.0013\pm .0001$ &             $.0024\pm .0001$ &             $.670\pm .002$ &             $1.277\pm .009$ \\
                                                   & $\textsc{NC}$ &             $.0032\pm .0001$ &             $.8250\pm .0176$ &             $.0017\pm .0001$ &             $.0039\pm .0002$ &             $.670\pm .002$ &             $1.502\pm .036$ \\
  
\midrule
\multirow{8}{*}{\rotatebox{90}{\texttt{tissue}}} & \cellcolor{blue!10}$\textsc{LN}$ & \cellcolor{blue!10}   $\mathbf{.0144\pm .0012}$ & \cellcolor{blue!10}            $.7293\pm .0359$ & \cellcolor{blue!10}            $.0050\pm .0011$ & \cellcolor{blue!10}            $.0100\pm .0015$ & \cellcolor{blue!10}            $.630\pm .001$ & \cellcolor{blue!10}   $\mathbf{1.012\pm .003}$ \\
                                                   & $\textsc{DC}$ &             $.0276\pm .0013$ &             $.9638\pm .0104$ & $\underline{.0021\pm .0005}$ &    $\mathbf{.0062\pm .0007}$ &             $.617\pm .003$ &             $1.052\pm .009$ \\
                                                   & $\textsc{KO}$ &             $.0177\pm .0006$ &    $\mathbf{.6913\pm .0349}$ &             $.0047\pm .0006$ &             $.0144\pm .0006$ & $\underline{.632\pm .001}$ &             $1.027\pm .002$ \\
                                                   & $\textsc{KC}$ & $\underline{.0165\pm .0008}$ & $\underline{.7179\pm .0250}$ &             $.0043\pm .0006$ &             $.0126\pm .0010$ &    $\mathbf{.632\pm .001}$ & $\underline{1.021\pm .002}$ \\
                                                   & $\textsc{PS}$ &             $.0418\pm .0018$ &             $.9521\pm .0043$ &    $\mathbf{.0013\pm .0001}$ &             $.0135\pm .0012$ &             $.603\pm .008$ &             $1.180\pm .008$ \\
                                                   & $\textsc{IR}$ &             $.0360\pm .0024$ &             $.9605\pm .0101$ &             $.0026\pm .0004$ & $\underline{.0095\pm .0005}$ &             $.603\pm .008$ &             $1.096\pm .016$ \\
                                                   & $\textsc{TS}$ &             $.0413\pm .0034$ &             $.9620\pm .0107$ &             $.0127\pm .0025$ &             $.0196\pm .0036$ &             $.603\pm .008$ &             $1.112\pm .023$ \\
                                                   & $\textsc{NC}$ &             $.0768\pm .0019$ &             $.9695\pm .0127$ &             $.0308\pm .0016$ &             $.0725\pm .0020$ &             $.603\pm .008$ &             $2.100\pm .101$ \\
\bottomrule
\end{tabular}
    \caption{Results for all experiments over five seeds.}
    \label{tab:ALLRES}
\end{table}

\cref{tab:ALLRES} reports all the results for \textbf{Q1}, \textbf{Q2} and \textbf{Q3}.

\end{document}